\documentclass[a4paper, conference]{IEEEtran}
\IEEEoverridecommandlockouts
\usepackage{amsmath,amssymb,amsfonts}
\usepackage{algorithmic}
\usepackage{graphicx}
\usepackage{textcomp}
\usepackage{xcolor}
\usepackage{graphicx}
\usepackage{subcaption}
\usepackage{bm}

\def\BibTeX{{\rm B\kern-.05em{\sc i\kern-.025em b}\kern-.08em
    T\kern-.1667em\lower.7ex\hbox{E}\kern-.125emX}}

\usepackage[backend=biber,
            style=numeric,
            abbreviate=false,
            dateabbrev=false,
            urldate=long,
            alldates=long]{biblatex}

\DefineBibliographyExtras{english}{%
}

\makeatletter

\def\ps@IEEEtitlepagestyle{%
  \def\@oddfoot{\mycopyrightnotice}%
  \def\@evenfoot{}%
}
\def\mycopyrightnotice{%
  {\footnotesize 978-1-6654-6623-3/22/\$31.00~\copyright~2022 IEEE\hfill}
  \gdef\mycopyrightnotice{}
}

\begin{document}

\title{EGAIN: Extended GAn INversion}

\author{\IEEEauthorblockN{1\textsuperscript{st} Wassim Kabbani}
\IEEEauthorblockA{\textit{Applied Math. and Comp. Science} \\
\textit{Technical University of Denmark}\\
Copenhagen, Denmark \\
se.wassim.kabbani@gmail.com}
\and
\IEEEauthorblockN{2\textsuperscript{nd} Marcel Grimmer}
\IEEEauthorblockA{\textit{Norwegian Biometrics Laboratory} \\
\textit{Norw. University of Science and Technology}\\
Gj{\o}vik, Norway \\
marceg@ntnu.no}
\and
\IEEEauthorblockN{3\textsuperscript{rd} Christoph Busch}
\IEEEauthorblockA{\textit{Norwegian Biometrics Laboratory} \\
\textit{Norw. University of Science and Technology}\\
Gj{\o}vik, Norway \\
christoph.busch@ntnu.no}
}

\maketitle

\begin{abstract}

Generative Adversarial Networks (GANs) have witnessed significant advances in recent years, generating increasingly higher quality images, which are non-distinguishable from real ones. Recent GANs have proven to encode features in a disentangled latent space, enabling precise control over various semantic attributes of the generated facial images such as pose, illumination, or gender. GAN inversion, which is projecting images into the latent space of a GAN, opens the door for the manipulation of facial semantics of real face images. This is useful for numerous applications such as evaluating the performance of face recognition systems. In this work, EGAIN, an architecture for constructing GAN inversion models, is presented. This architecture explicitly addresses some of the shortcomings in previous GAN inversion models. A specific model with the same name, \textit{egain}, based on this architecture is also proposed, demonstrating superior reconstruction quality over state-of-the-art models, and illustrating the validity of the EGAIN architecture.

\end{abstract}


\section{Introduction}

Face recognition has become more popular with increasing application scenarios, ranging from border control, criminal investigations and entertainment. Specifically, face recognition refers to the automated recognition of individuals based on their facial characteristics. In this context, the main challenge for these systems is to be robust against intra-identity variations while being sensitive to inter-identity variations. While the revolution of deep learning-based face recognition systems significantly improved the generalizability, it remains challenging to verify probe images with extreme factors of variation (e.g. long-term ageing).

One possible solution for reducing the intra-identity variation is to use generative adversarial networks (GANs) to re-balance single factors of variations with extreme expressions. With the introduction of StyleGAN2 \cite{karras2020analyzing} and SWAGAN \cite{gal2021swagan}, the generation of facial images has reached a state where fully-synthetic face images have photo-realistic quality and become indistinguishable from real ones. Moreover, the GAN-internal data representations (latent codes) encapsulate rich semantic information from the original face images. The latent codes form a space (latent space) with a disentangled representation of facial attributes. A latent space is called fully disentangled if a single variation factor can be edited without changing other facial attributes (e.g. ageing without adding eyeglasses). 
The continuous improvement of the disentanglement properties enables the inversion of real face images to the latent space of a pre-trained GAN. In other words, GAN inversion refers to the task of finding the corresponding latent code $w$ of a given face image $x$, which when passed to the generator, generates a reconstructed face image $\hat{y}$ close to the original. Once a real face image is inverted to the latent space, the latent code can be edited to manipulate single facial attribute known to have a negative impact on the face recognition accuracy, such as ageing effects or head pose rotations.

The main contribution of this work is an extended GAN inversion architecture (EGAIN) designed to give a reference for constructing GAN inversion models that produce latent codes, from which a GAN generator can produce high fidelity reconstructed images. Operating within EGAIN, this work also proposes one GAN inversion model \textit{egain} that realizes this architecture and demonstrates its effectiveness.

This paper is structured as follows: First, Section \ref{sec:related} highlights some of the related works. Then, Section \ref{sec:proposed} explains the main concepts of the EGAIN architecture, followed by an introduction to the proposed \textit{egain} model. Next, Section \ref{sec:settings} describes the datasets and metrics used to evaluate the performance of \textit{egain}. The experimental results are presented in Section \ref{sec:experiments}, analysing the quality of the reconstructed face images from the perspective of image similarity and biometric quality. Furthermore, the results are compared to related state-of-the-art GAN inversion and reconstruction techniques. Finally, Section \ref{sec:conlusion} summarises the main findings and outlines potential directions for future research.

\section{Related Work}
\label{sec:related}

\textbf{Generative Adversarial Networks (GANs)}. GANs were first introduced by Ian Goodfellow \cite{goodfellow2014generative} where he described an architecture for training a generative model. The architecture comprises two modules: a generator and a discriminator, both built using neural networks. The main goal of the generator is to learn how to map random noise to a target domain (e.g. face images). At the same time, the discriminator becomes better at judging whether the generated samples are real or fake. Training a GAN model is a complex and unstable process. Therefore,  Radford et al.~\cite{radford2016unsupervised} introduced Deep Convolutional GANs (DCGANs), providing a more stable training routine. Many traditional GANs start the content generation in the feature space, learning a hierarchy of representations from object parts to scenes in both the generator and the discriminator. However, this produced images with low visual quality \cite{radford2016unsupervised} \cite{mirza2014conditional}. Instead, more recent GANs generate hierarchical content directly in the image space, generating images with higher resolutions and better visual quality. In 2018, Karras et al.~\cite{karras2018progressive} proposed a progressively growing generator and discriminator architecture, learning facial attributes from coarse-to-fine details.

Inspired by the style transfer literature, which enables changing the style of an image while preserving its content \cite{huang2017arbitrary}, Karras et al. introduced a style-based generator architecture \cite{karras2019stylebased}. The proposed StyleGAN \cite{karras2019stylebased} generator automatically learns the separation between high-level and low-level facial attributes, allowing for mixing the styles of multiple identities. Later, StyleGAN2 was published by the same authors with an improved network architecture that yields better image quality and better disentanglement of the latent space. Recently, StyleGAN3 \cite{karras2021aliasfree} was introduced to enable smooth transitions  
in the latent space, optimizing the usage for video and animation creation applications.

Previous research has shown that the generation quality of GANs suffers when handling high-frequency content. A study by Chen et al. \cite{chen2020ssdgan} observed the issue of missing high-frequency details in the discriminator of GANs. Consequently, the generator lacks the incentive to learn the high-frequency content of the training data, resulting in a significant spectrum discrepancy between the generated and real samples. Moreover, Dzanic et al. \cite{dzanic2020fourier} analyzed the high-frequency Fourier modes of real and deep network-generated images. They demonstrate that deep network-generated images share an observable and systematic shortcoming in replicating data with high-frequency details. Further, Durall et al. \cite{durall2020watch} show that common up-sampling methods, which are frequently used in GAN architectures, cause the inability of such models to learn the spectral distribution of the training dataset.

Style-based WAvelet-driven Generative Model (SWAGAN), introduced by Gal et al. \cite{gal2021swagan}, aims to address this problem by integrating discrete wavelet transformations into the network architecture. This strategy enables the generator and discriminator to focus more on high-frequency details.

\textbf{GAN Inversion}. The concept of GAN inversion aims to invert a source  image into the latent space of a pre-trained GAN model, such that the image can be reconstructed from the inverted code by the generator. The latent code must fulfill two goals: (1) reconstruct the source image with high fidelity and (2) facilitate downstream tasks such as attribute editing \cite{xia2021gan}. GAN inversion paves the way for real face images to be edited by manipulating their latent codes in the latent space.

In a study by Zhu et al. \cite{zhu2018generative}, the authors propose to learn the natural image manifold from real image data using a generative adversarial network. After performing editing operations, the model automatically adjusts the latent code to force it on the learned manifold, keeping the edits as realistic as possible. In another study \cite{pan2020exploiting}, the authors exploit the learned prior distribution of a GAN to allow the restoration of missing image semantics such as colour, patch and resolution. It also enables diverse image manipulations, including image morphing and category transfer. 

Motivated by the ability to apply latent space manipulations to real images for applications such as face age progression \cite{marcel2021age}, facial frontalization, inpainting and super-resolution \cite{richardson2021encoding}, the task of GAN inversion has received much attention. Generally, three approaches for GAN inversion can be identified: optimization-based, learning-based and hybrid \cite{xia2021gan}. Optimization-based methods are based on an optimization algorithm that optimizes the latent code of a given image to minimize the reconstruction error between the original image and the reconstructed image. These techniques provide the highest reconstruction quality, but with the cost of long processing time \cite{abdal2019image2stylegan} \cite{pinnimty2020transforming}. Learning-based methods train an encoder once to learn the mapping between image and latent space. While they are significantly faster, they suffer from inferior image quality \cite{pidhorskyi2020adversarial} \cite{richardson2021encoding} \cite{tov2021designing}. Hybrid methods attempt to balance the trade-offs between the previous two strategies, typically using an encoder only for the initialization of a latent code to start the optimization process \cite{zhu2020indomain} \cite{chai2021ensembling}.

\section{Proposed Method}
\label{sec:proposed}

The typical learning-based GAN inversion architecture comprises two main components: an encoder $E$ and a decoder $G$, where $G$ is the generator part of a pre-trained GAN model. The encoder $E$ is trained such that given a source image $x$, it outputs a latent code $w$, which, when passed to the generator $G$, generates a reconstructed image $\hat{y}$ that is as similar as possible to the source image $x$.

These GAN inversion architectures suffer typically from two problems. First, the information bottleneck theory illustrates that deep models primarily learn general patterns from the training data while ignoring finer details \cite{tishby2015deep}. This compression property also applies in encoders trained for GAN inversion, leading to reconstructed images with little high-frequency information. Second, the inherent editability-distortion trade-off in the latent space of GAN models \cite{tov2021designing} requires the balance of both factors. When designing an encoder optimized for reducing image distortion, the corresponding latent codes end up in areas of the latent space far away from the learned face manifold, reducing the editability. In contrast, when optimizing an encoder on editability, the resulting latent codes will be located on the face manifold, with the cost of missing details from the original image. 

The above-described issues are addressed in this work by proposing an Extended GAn INversion architecture (EGAIN). Instead of relying on a single encoder, the basic architecture is extended with another branch for calculating the information delta between the source image $x$ and the initially reconstructed image $\hat{y}_0$. This delta is then passed to the generator to guide the generation of higher fidelity face images, capturing most of the details of the source image. EGAIN proposes a novel architecture that comprises five distinct components:

\textbf{Basic Encoder $E_b$:} This component is similar to the encoder present in the classic GAN inversion architecture. Given a source image $x$, it outputs an initial low-dimensional latent code $w_b$ corresponding to an initial low-fidelity reconstructed image $\hat{y}_0$.

\textbf{Delta Encoder $E_{\Delta}$:} This component encodes the computed difference $\Delta_0$ between the source image $x$ and the initial low-fidelity reconstructed image $\hat{y}_0$. Depending on the choice of the fusion scheme, this delta encoder outputs a latent code $w_d$ with the same size as $w_b$, or a set of latent feature maps $M_d$.

\begin{figure}[h]
    \centering
    \includegraphics[width=0.50\textwidth]{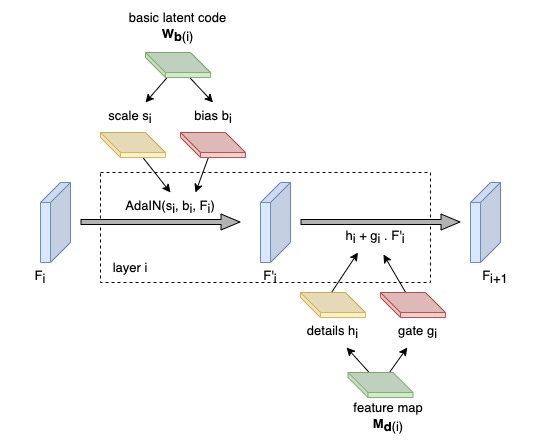}
    \caption{Internal Fusion Scheme - In layer $i$ of the generator $G$, feature map $M_{d(i)}$ is embedded to a gate map $g_i$ and a high-frequency details map $h_i$. These are combined with the low-fidelity features obtained from $w_{b(i)}$ to produce high-fidelity feature maps $F{i+1}$.}
    \label{fig:arch-fusion}
\end{figure}

\begin{figure}[h] 
  \begin{subfigure}[b]{1.0\linewidth}
    \centering
    \includegraphics[width=0.90\linewidth]{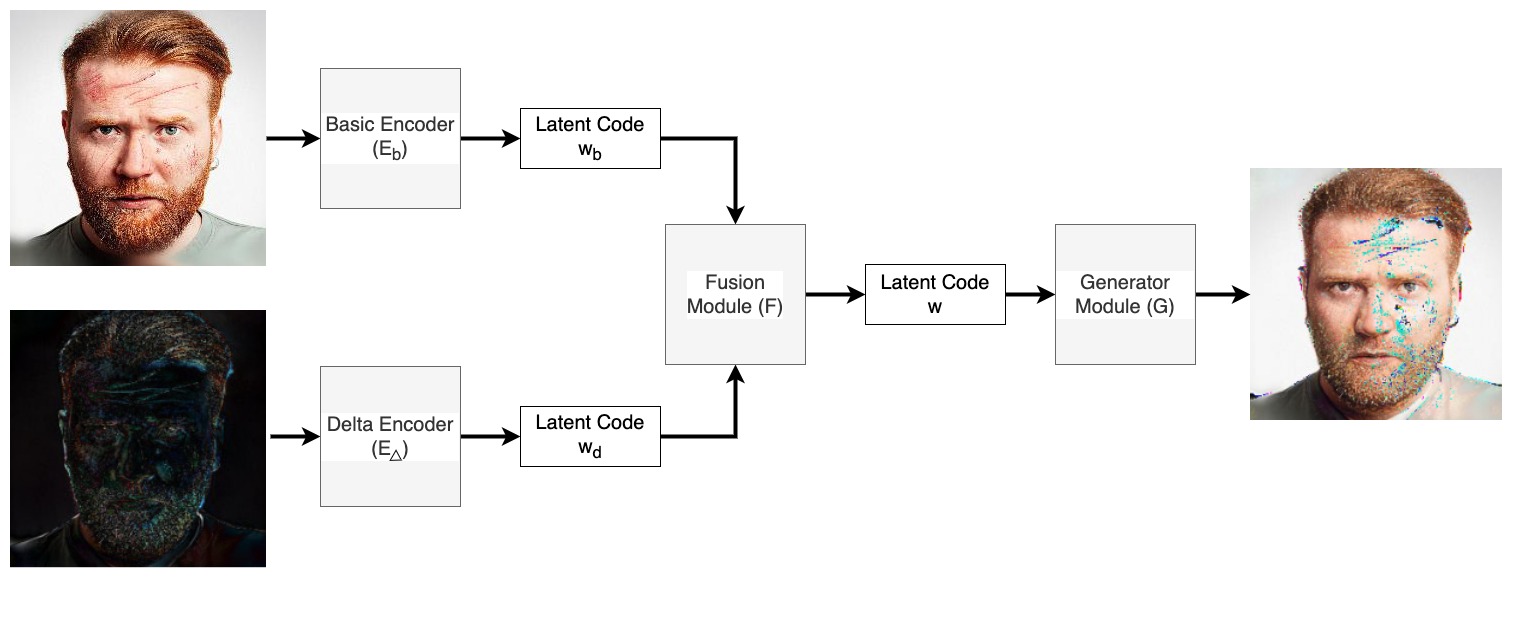}
    \caption{External Fusion} 
    \label{fig:arch-external}
  \end{subfigure}
  \begin{subfigure}[b]{1.0\linewidth}
    \centering
    \includegraphics[width=0.90\linewidth]{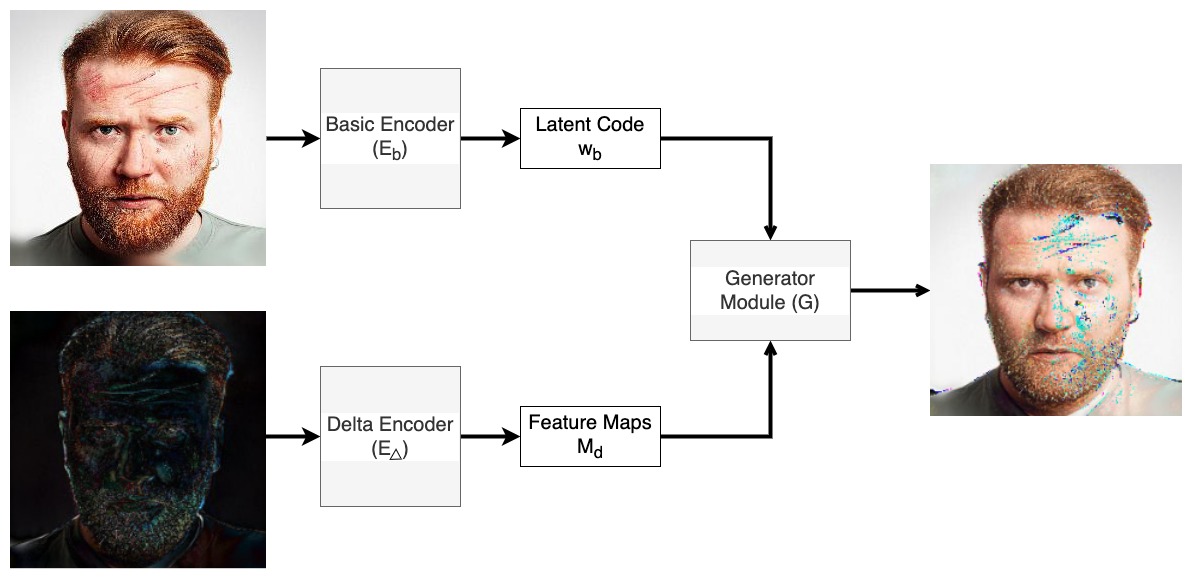}
    \caption{Internal Fusion} 
    \label{fig:arch-internal}
  \end{subfigure}
  \caption{The source image $x$ is shown in the upper left corner. $\Delta$ between $x$ and $\hat{y}_0$ is shown in lower left corner and the reconstructed image $\hat{y}$ is shown on the right  (an illustration image is used to highlight the concept of the difference $\Delta$)}
  \label{fig:qualitative} 
\end{figure}

\textbf{Fusion Module $F$:} This component is responsible for fusing the outputs of the basic encoder and the delta encoder. Conceptually, the fusion can be external (outside the generator) or internal (inside the generator). In the external fusion case, illustrated in figure \ref{fig:arch-external}, $w_b$ and $w_d$ are combined before passing the result $w$ to the generator $G$ to obtain the reconstructed image $\hat{y}$. In the internal fusion case, illustrated in figure \ref{fig:arch-internal}, the fusion is conducted inside the generator. To achieve this, the internal fusion scheme of Wang et al. \cite{wang2021HFGI} is adopted and applied with SWAGAN. The internal fusion scheme, illustrated in figure \ref{fig:arch-fusion}, exploits the architecture of the generator in such a way that the delta code $w_b$ and the delta feature map $M_d$ are mapped to the internal generator layers.

\textbf{Generator Module $G$:} This decoder component generates the reconstructed image. It is responsible for generating two different reconstructed images: The first one is the initial reconstructed image $\hat{y}_0$, given the initial low-dimensional latent code $w_b$. The second one is the final image generated by the model after applying the external or internal fusion scheme. 

\textbf{Delta Calculator $C_{\Delta}$:} This component is responsible for computing the information delta $\Delta$ between the source image and the reconstructed image. $\Delta$ represents the lost high-frequency image-specific details that can be used to improve the reconstruction quality.

This clear identification of components and separation of responsibilities constitute a modular architecture. It allows EGAIN to be flexible, changing any network component or pre-trained models independent of other modules. EGAIN is also training-scheme independent and can be used with iterative training and inference schemes where the final reconstructed image results from an iterative process that repeats the entire flow through the architecture multiple times. Alternatively, it can be used with a simple single-pass training and inference scheme. Moreover, EGAIN explicitly addresses the editability-distortion trade-off by capturing and conveying the image-specific details separately. The delta information relieves the basic encoder from the necessity of producing less editable latent codes in an attempt to reduce distortion. Rather, a basic encoder optimized to produce highly editable latent codes can be used, leaving the job of capturing and conveying image details to the delta component.

EGAIN describes a generic architecture for GAN inversion rather than a specific model implementation. Operating within EGAIN, a specific model, \textit{egain}, is proposed based on this architecture using the internal fusion variant. The model comprises the five EGAIN components, i.e. basic encoder, delta encoder, fusion module, generator module and delta calculator, and the individual composition defines the uniqueness of \textit{egain}. For the basic encoder $E_b$, a pre-trained model of the \textit{pSp} encoder is used. \textit{pSp} is a generic image to image translation framework presented by Richardson et al. \cite{richardson2021encoding}. It introduces a novel encoder network that embeds images into StyleGAN's latent space. For the delta calculator $C_{\Delta}$, a subtraction operation is used that calculates the difference between the source image $x$ and the initially reconstructed image $\hat{y}_0$. The delta encoder $E_{\Delta}$ uses a custom-built convolutional neural network to extract the feature maps $M_d$ from $\Delta_0$. The fusion process is done inside the generator using the internal fusion scheme shown in figure \ref{fig:arch-fusion}. The fusion module implicitly fuses the feature maps $M_d$ and the basic latent code $w_b$ inside the generator $G$. Furthermore, a custom-trained SWAGAN generator is used, replacing the StyleGAN generator due to the better capability of preserving high-frequency details. The custom-trained SWAGAN model was trained for $25$ days on a single \textit{Tesla V100 32 GB} GPU.

Training and inference schemes follow simple single-pass processes, where the training is performed as follows: for each training iteration and given a batch of images $\bm{x}$, (1) encode $\bm{x}$ with the basic encoder $E_b$ and obtain a batch of latent codes $\bm{w_b}$ (2) normalize with respect to the center of an average face by adding a batch of repeated average latent codes $\bm{\Bar{w}}$ (3) using the updated latent codes, generate the initial reconstructed images $\bm{\hat{y}_0}$ with $G$ (4) compute the difference $\bm{\Delta_0}$ between the original images $\bm{x}$ and $\bm{\hat{y}_0}$ (5) extract the features of the difference $\bm{\Delta_0}$ using the delta encoder $E_{\Delta}$ (6) use the generator $G$ again to generate the final reconstructed images $\hat{\bm{y}}$. Once this sequence is done, the reconstruction loss between $\bm{x}$ and $\hat{\bm{y}}$ is computed. The reconstruction loss, summarized in equation \ref{eq_total_loss}, is a weighted combination of individual loss functions and regularization terms. EGAIN contains measures to quantify the distortion in the reconstructed image such as Mean Square Error (L2), Learned Perceptual Image Patch Similarity (LPIPS), Face Identity Loss (ID) and Avg-Regularization $L_{avg-reg}$ which encourages the encoder to output latent style codes closer to the average latent code. Furthermore, other measures to assess the editability of the inverted latent codes such as Delta-Regularization $L_{d-reg}$ and W-Regularization $L_{w-reg}$ both suggested by Tov et al. \cite{tov2021designing} and designed to encourage the editability of the style codes.

\begin{equation}
\begin{split}
\label{eq_total_loss}
L(x) = \lambda_{d-reg} L_{d-reg}(x) + \lambda_{w-reg} L_{w-reg}(x) \\
+ \lambda_{l2} L_2(x) + \lambda_{lpips} L_{LPIPS}(x) \\ 
+ \lambda_{id} L_{ID}(x) + \lambda_{avg-reg} L_{avg-reg}(x)
\end{split}
\end{equation}

\section{Experimental Settings}
\label{sec:settings}

\textbf{Datasets.} For training \textit{egain}, the FFHQ dataset is used \cite{karras2019stylebased}. Afterwards, a custom-made evaluation dataset comprised of $1,510$ face images is collected, with the vast majority of images extracted randomly from the CelebAMask-HQ dataset \cite{lee2020celeba}. The CelebAMask-HQ dataset contains high-resolution 1024x1024 face images selected from the original CelebA dataset \cite{liu2015faceattributes} and is typically used in related works to benchmark the performance across multiple GAN inversion techniques \cite{tov2021designing} \cite{richardson2021encoding} \cite{wang2021HFGI} \cite{alaluf2021restyle}. This selection ensures consistent and compatible results with other works. In addition, $10$  face images are added to the dataset since they are featured in recent state-of-the-art works \cite{wang2021HFGI}.

\textbf{Metrics.} To evaluate the performance of the proposed model and prepare the comparison to state-of-the-art GAN inversion models, the following metrics are evaluated: 

\textbf{Face Identity (ID)} Since this work operates with face images, there is a particular interest in preserving the facial identity between the source and the reconstructed image. To this end, the pre-trained \textit{ArcFace} \cite{deng2019arcface} face recognition network is used to measure the identity loss after the reconstruction. Specifically, the cosine similarity between the feature embedding of the source image and the reconstructed image is computed.

\textbf{Structural Similarity (SSIM)} introduced by Wang et al. \cite{wang2004ssim} is used to measure the structural similarity between the source and the reconstructed images. It extracts three key features: luminance, contrast and structure. The SSIM metric is designed to output scores between $-1$ and $+1$. The score $+1$ indicates that the two given images are identical, while a score of $-1$ indicates a high structural difference. In this work, this range is normalized to an interval between $0$ and $1$.

\textbf{Spatial Correlation Coefficient (SCC)} introduced by Zhou et al. \cite{zhou1998scc} is another metric to measure the similarity between the source and the reconstructed images. It uses the correlation coefficients between the high-pass filtered source image and the high-pass filtered reconstructed image as an index of the spatial quality. This measurement is based on spatial information concentrated in the high-frequency domain. The higher the correlation in the high-frequency domain, the more spatial information from the source image is incorporated into the reconstructed image.

\textbf{Visual Information Fidelity (VIF)} introduced by Sheikh et al. \cite{Sheikh2006vifp} is an image quality assessment method. Unlike traditional image quality assessment algorithms, which predict visual quality by comparing a distorted image against a reference image, typically by modeling the human visual system or by using arbitrary signal fidelity criteria. This method proposes an information fidelity criterion that quantifies the Shannon information shared between the reference and the distorted image, relative to the information contained in the reference image itself. Beside a human visual system model, it uses Natural Scene Statistics (NSS) modeling and an image degradation model.

\textbf{Face Recognition Quality Assessment} Recognizing faces in the wild is a challenging task for face recognition systems \cite{rowden2018quality}. Thus, face quality assessment algorithms (FQAAs) are employed before storing face images as biometric references to ensure a high recognition accuracy for later comparisons to biometric probe samples. The quality score should therefore be indicative of the face recognition performance. In this work, \textit{MagFace} is used to predict the face recognition utility of the reconstructed images, comparing it to the behaviour of the source face images. \textit{MagFace} was introduced by Meng et al. \cite{meng2021magface}, who introduced a universal face representation applicable for face recognition and face quality assessment. In essence, MagFace introduces a category of losses for learning universal face embeddings whose feature magnitudes act as preditors of their biometric quality.

\section{Experimental Results}
\label{sec:experiments}

The results of the metrics, presented in table \ref{tbl:quantitative}, show a clear superiority of \textit{egain} over all the state-of-the-art models and by a wide margin. The Spatial Correlation Coefficient (SCC) results are specifically interesting since they represent the performance analysis in the high-frequency domain. The SCCs indicate a clear performance advantage of \textit{egain} over other models. This observation is assumed to be based on SWAGAN, a network designed to preserve high-frequency information better than StyleGAN2. The results of the MagFace assessment do not show any significant performance differences between the models. They all perform equally well since the reconstructed images are still good enough in the context of face recognition applications.

\begin{table}
    \centering
    \begin{tabular}{c|c|c|c|c|c}
        Model & Face ID $\uparrow$  & SSIM $\uparrow$ & SCC $\uparrow$ & VIF $\uparrow$ & MagFace $\uparrow$ \\
        e4e & 0.51 & 0.50 & 0.03 & 0.40 & 25.89 \\
        pSp & 0.57 & 0.54 & 0.04 & 0.43 & 25.85 \\
        restyle(e4e) & 0.51 & 0.52 & 0.04 & 0.52 & 25.61 \\
        reStyle(pSp) & 0.66 & 0.58 & 0.06 & 0.62 & \textbf{25.97} \\
        hfgi & 0.69 & 0.62 & 0.05 & 0.58 & \textbf{25.97} \\
        egain & \textbf{0.85} & \textbf{0.81} & \textbf{0.18} & \textbf{0.79}  & 25.65 \\
    \end{tabular}
    \caption{Face image quality measures (higher scores are better)}
    \label{tbl:quantitative}
\end{table}

\begin{figure}[ht] 
  \begin{subfigure}[b]{0.5\linewidth}
    \centering
    \includegraphics[width=0.75\linewidth]{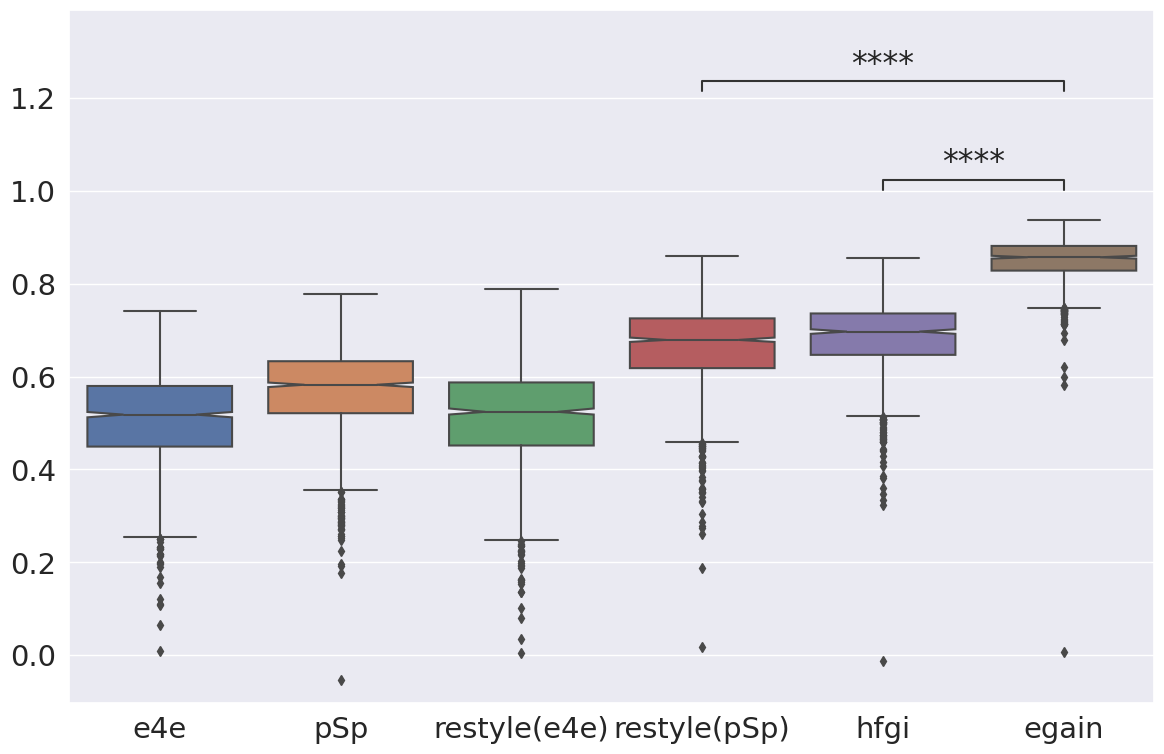} 
    \caption{Face ID} 
    \label{fig7:a} 
    \vspace{3ex}
  \end{subfigure}
  \begin{subfigure}[b]{0.5\linewidth}
    \centering
    \includegraphics[width=0.75\linewidth]{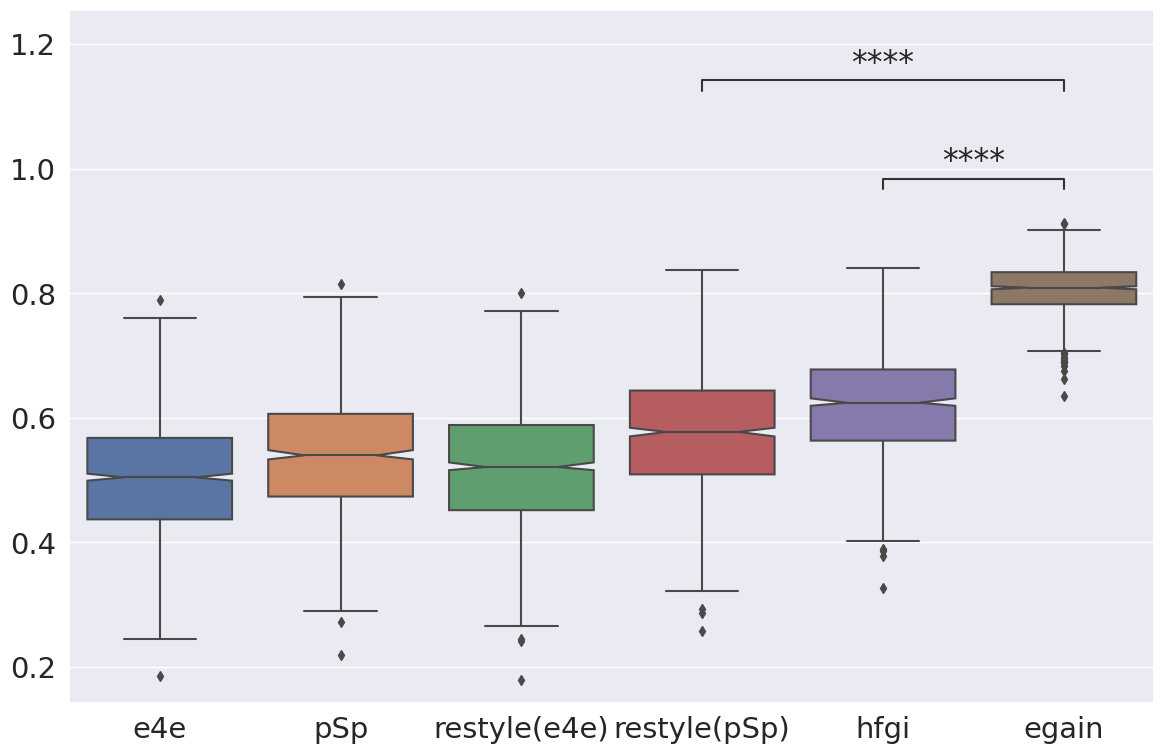} 
    \caption{SSIM} 
    \label{fig7:b} 
    \vspace{3ex}
  \end{subfigure} 
  \begin{subfigure}[b]{0.5\linewidth}
    \centering
    \includegraphics[width=0.75\linewidth]{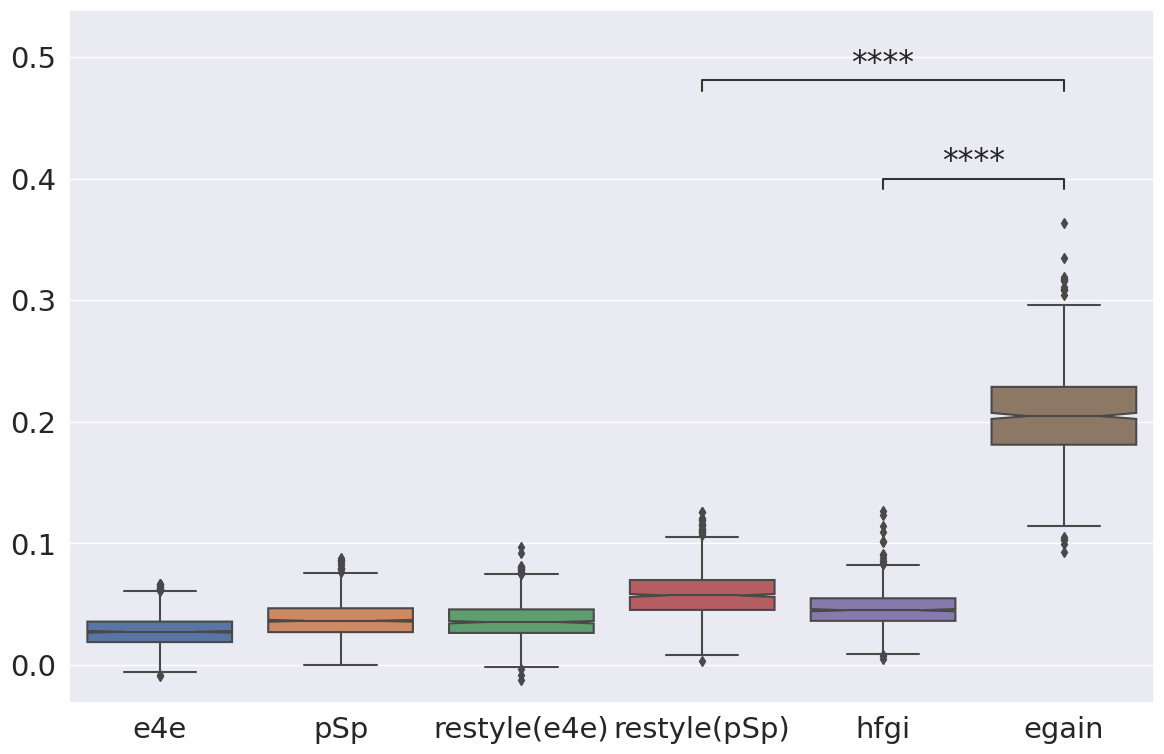} 
    \caption{SCC}
    \label{fig7:c} 
  \end{subfigure}
  \begin{subfigure}[b]{0.5\linewidth}
    \centering
    \includegraphics[width=0.75\linewidth]{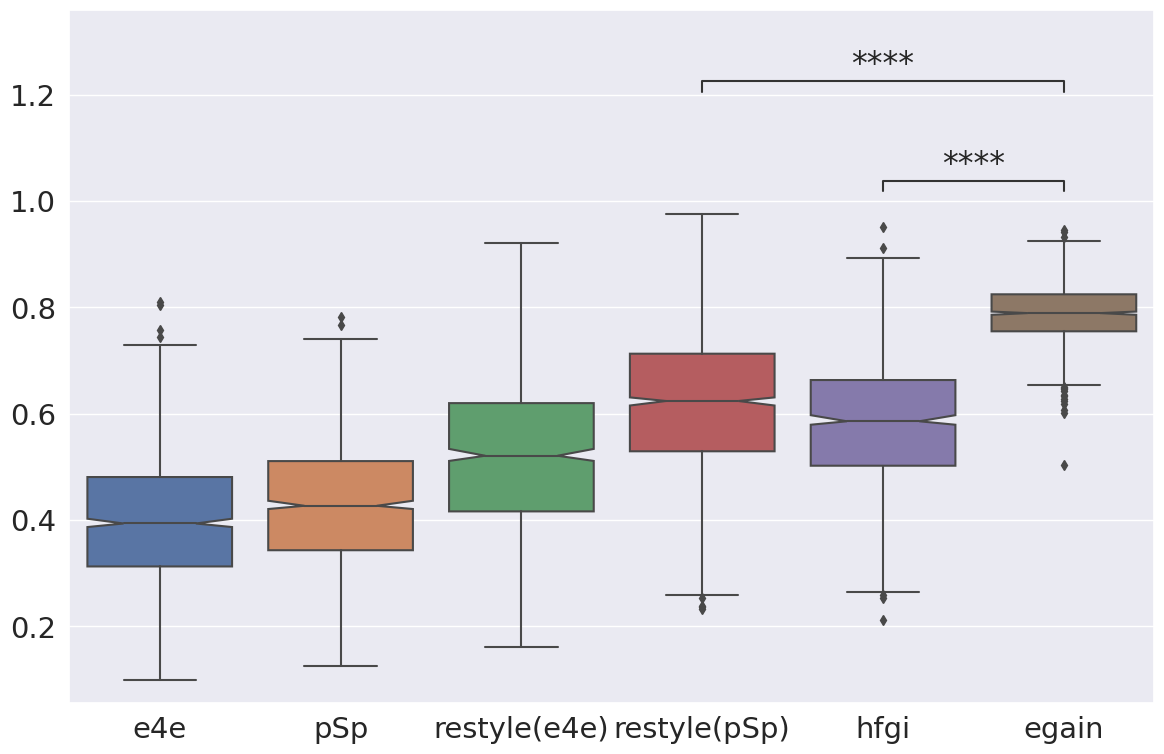} 
    \caption{VIF}
    \label{fig7:d} 
  \end{subfigure} 
  \caption{The results of the face image quality measures also indicate that the difference between the medians of the scores of \textit{egain} and the otherwise best performing models \textit{hfgi} and \textit{restyle(pSp)} is statistically significant.}
  \label{fig:qualitative} 
\end{figure}

The visualisation of reconstruction quality, illustrated in Figure \ref{fig:qualitative}, show that the proposed \textit{egain} model has superior performance over all the state-of-the-art models with regard to preserving the details of the source images during the reconstruction. In columns (c) and (d) of Figure \ref{fig:qualitative}, it can be seen that \textit{hfgi} and \textit{egain} are the only models able to reconstruct the hand of the woman and the watch in front of the man's eye. In contrast,  the other models unrealistically enlarged the neck area in column (c), and became confused about the watch, misinterpreting it as an extension of other facial features in column (d). Also, it can be noticed that minor details are better captured in the reconstructed images of the \textit{egain} model, such as the shape of the sunglasses, the colour of the fingernails, and the details of the watch. The differences in high-quality face images, such as depicted in column (a), are more subtle. However, it can also be noticed that the earrings appear only in the reconstructed images of models \textit{hfgi} and \textit{egain}.

\begin{figure}
\centering
\begin{tabular}{ccccc}
    
     &
    (a) &
    (b) &
    (c) &
    (d) \\
    
    \rotatebox{90}{Original} &
    \includegraphics[width=.18\linewidth]{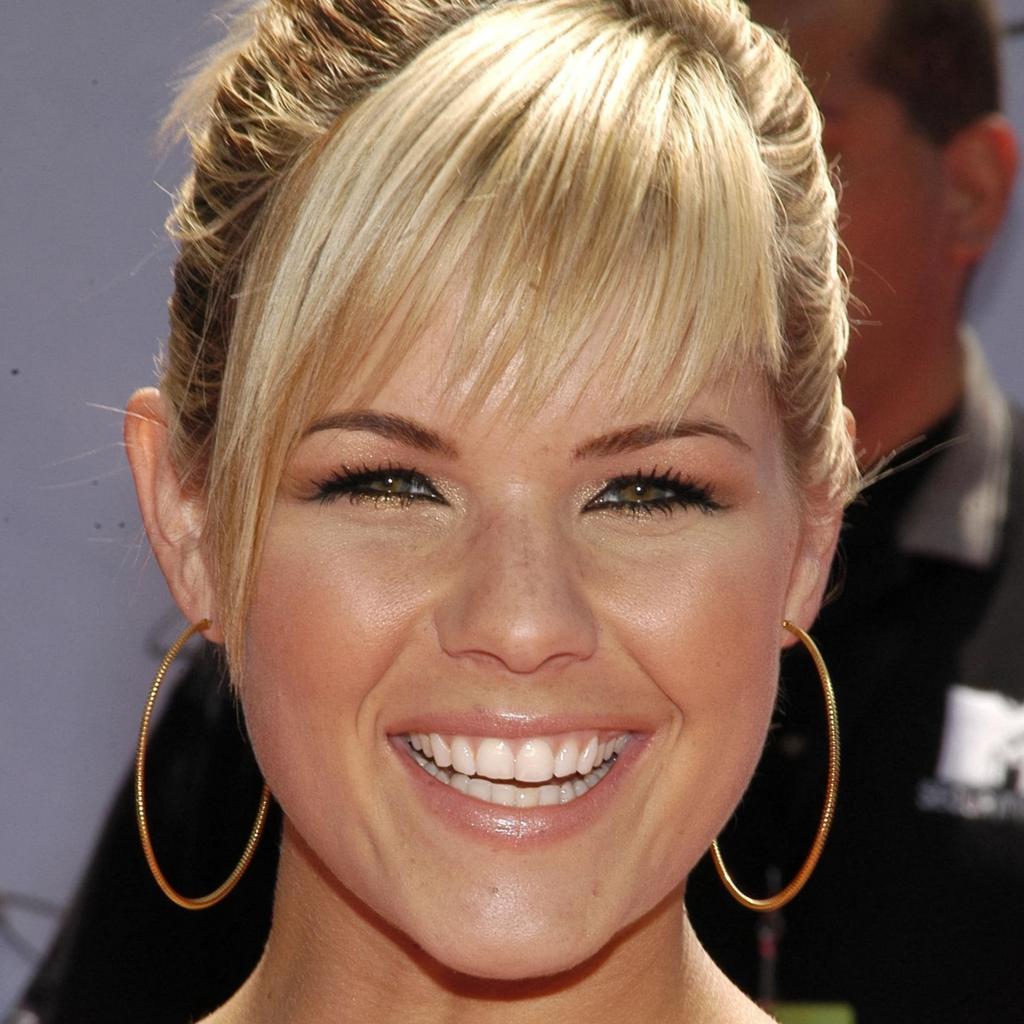} &
    \includegraphics[width=.18\linewidth]{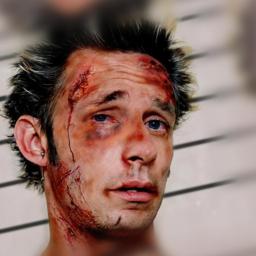} &
    \includegraphics[width=.18\linewidth]{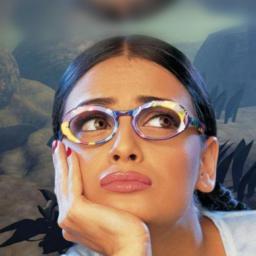} &
    \includegraphics[width=.18\linewidth]{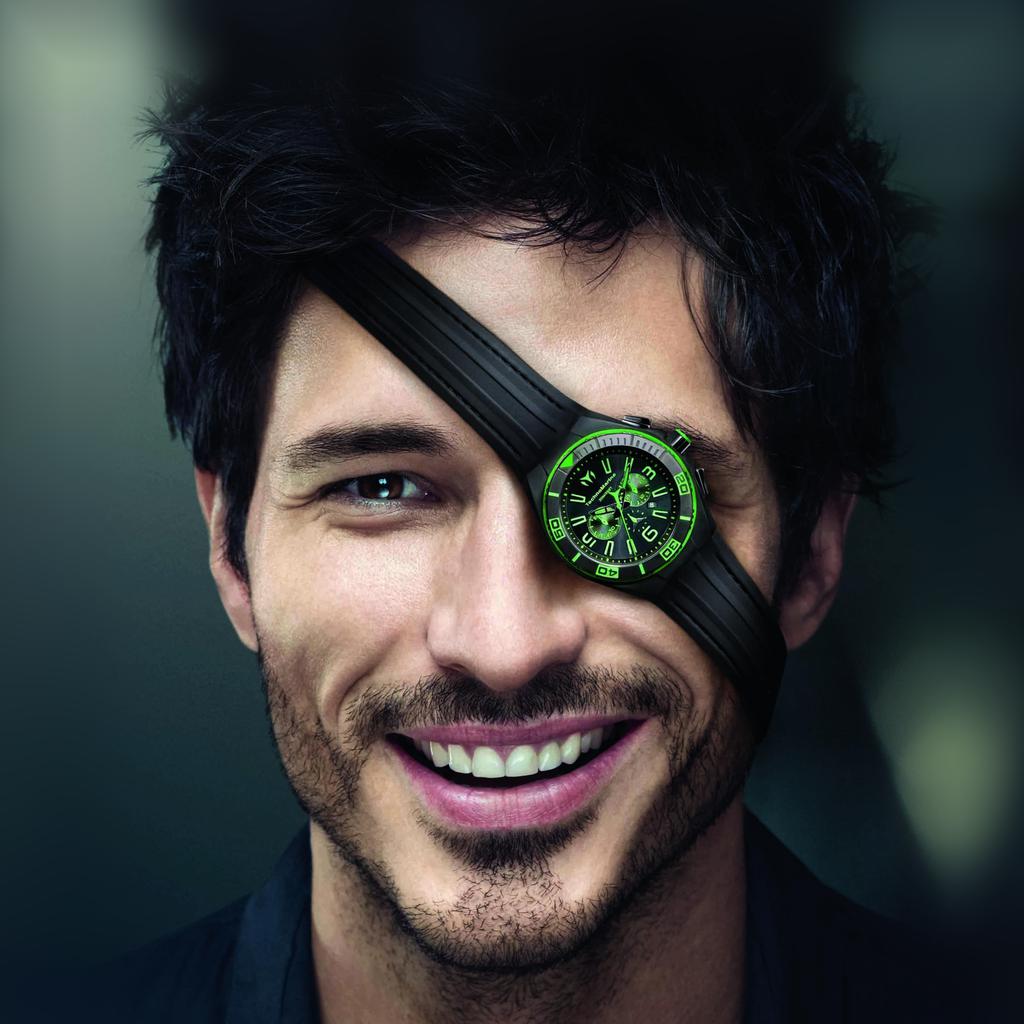} \\
    
    \rotatebox{90}{e4e} &
    \includegraphics[width=.18\linewidth]{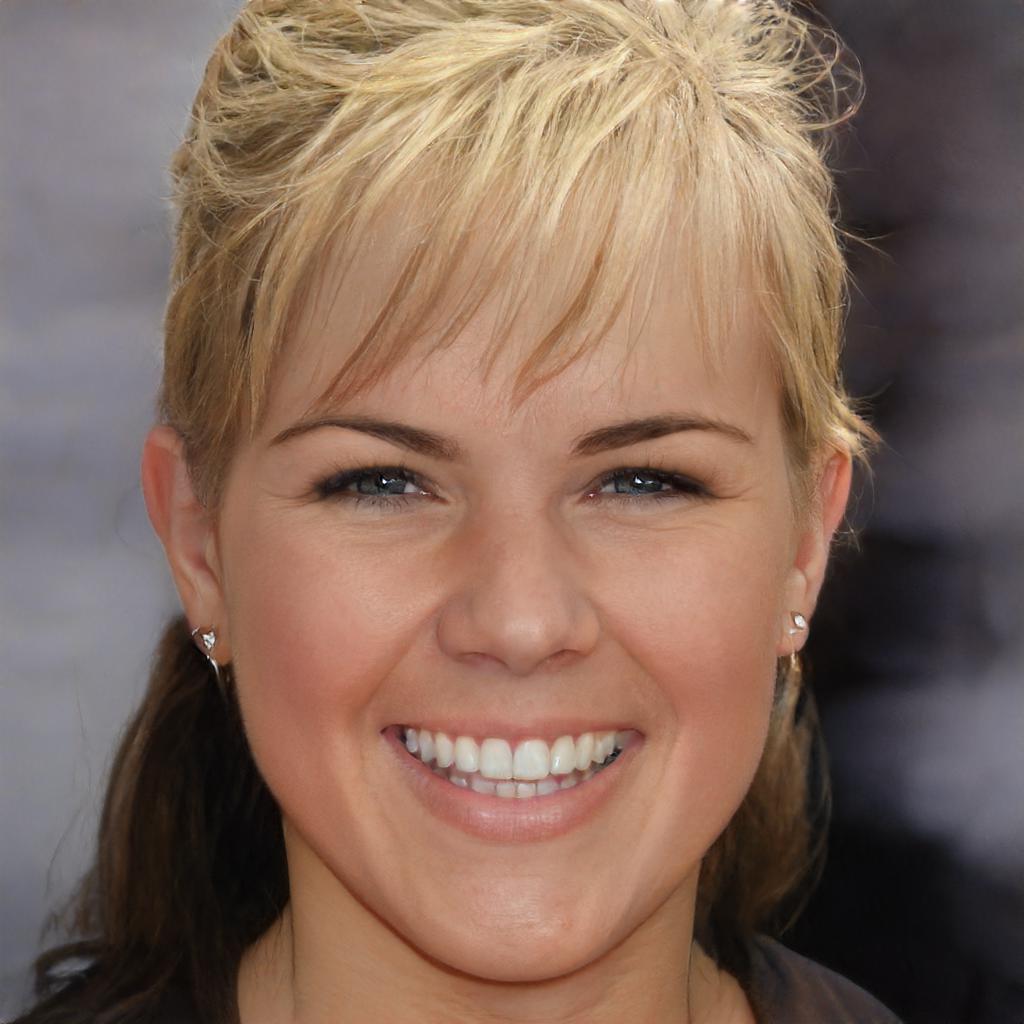} &
    \includegraphics[width=.18\linewidth]{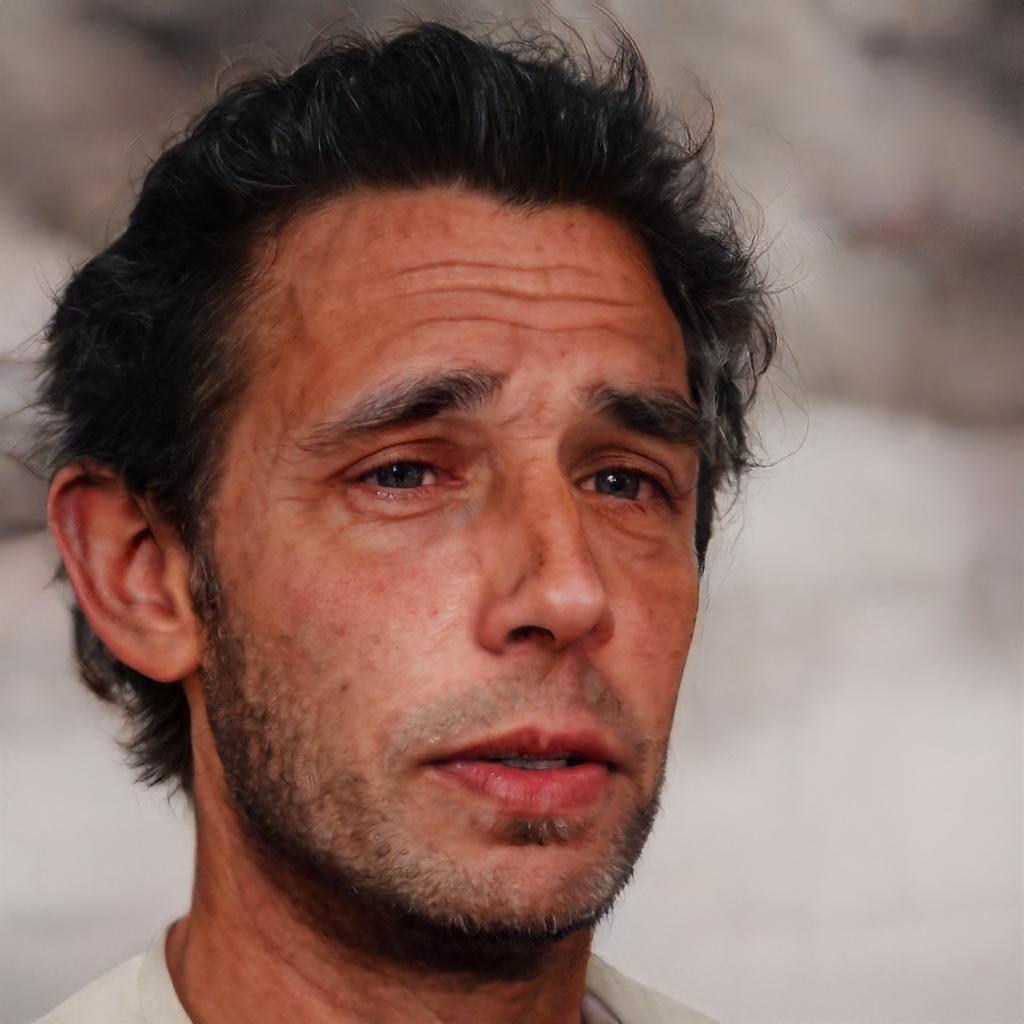} &
    \includegraphics[width=.18\linewidth]{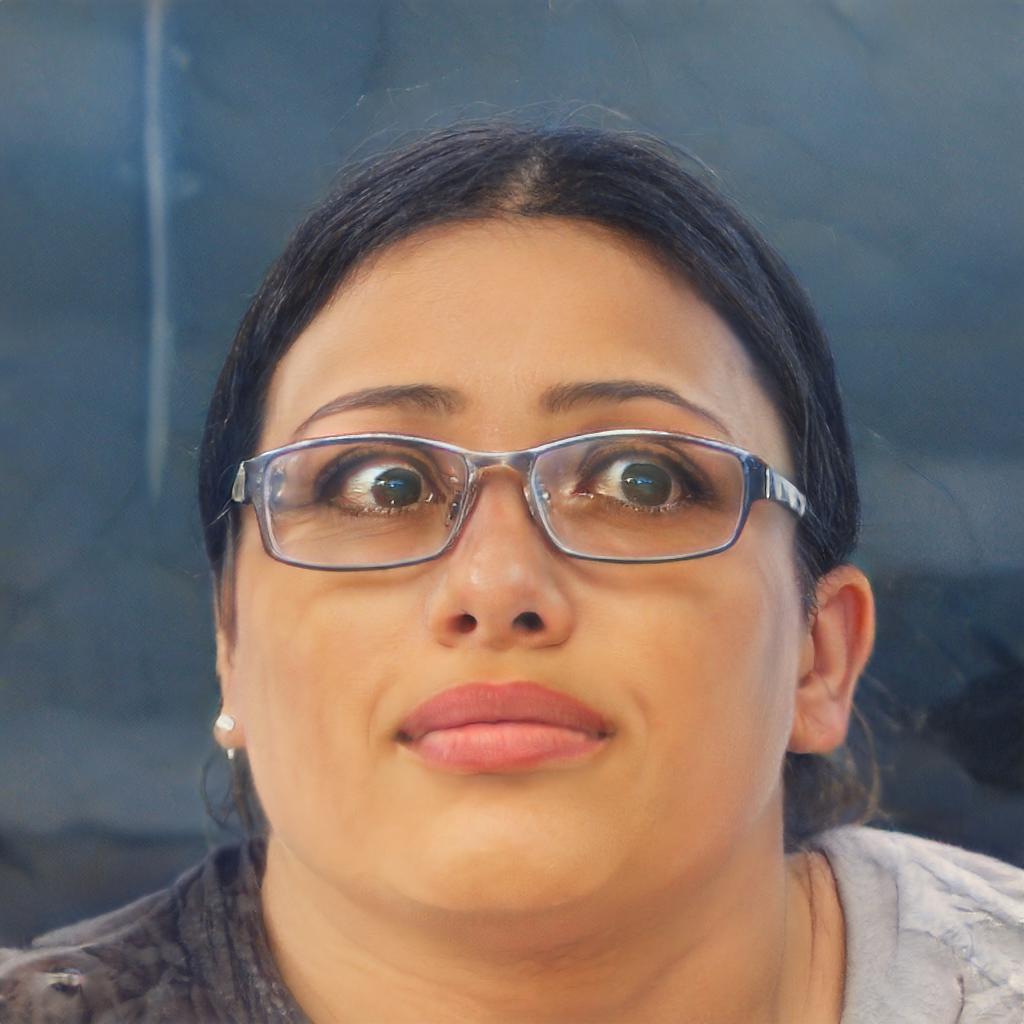} &
    \includegraphics[width=.18\linewidth]{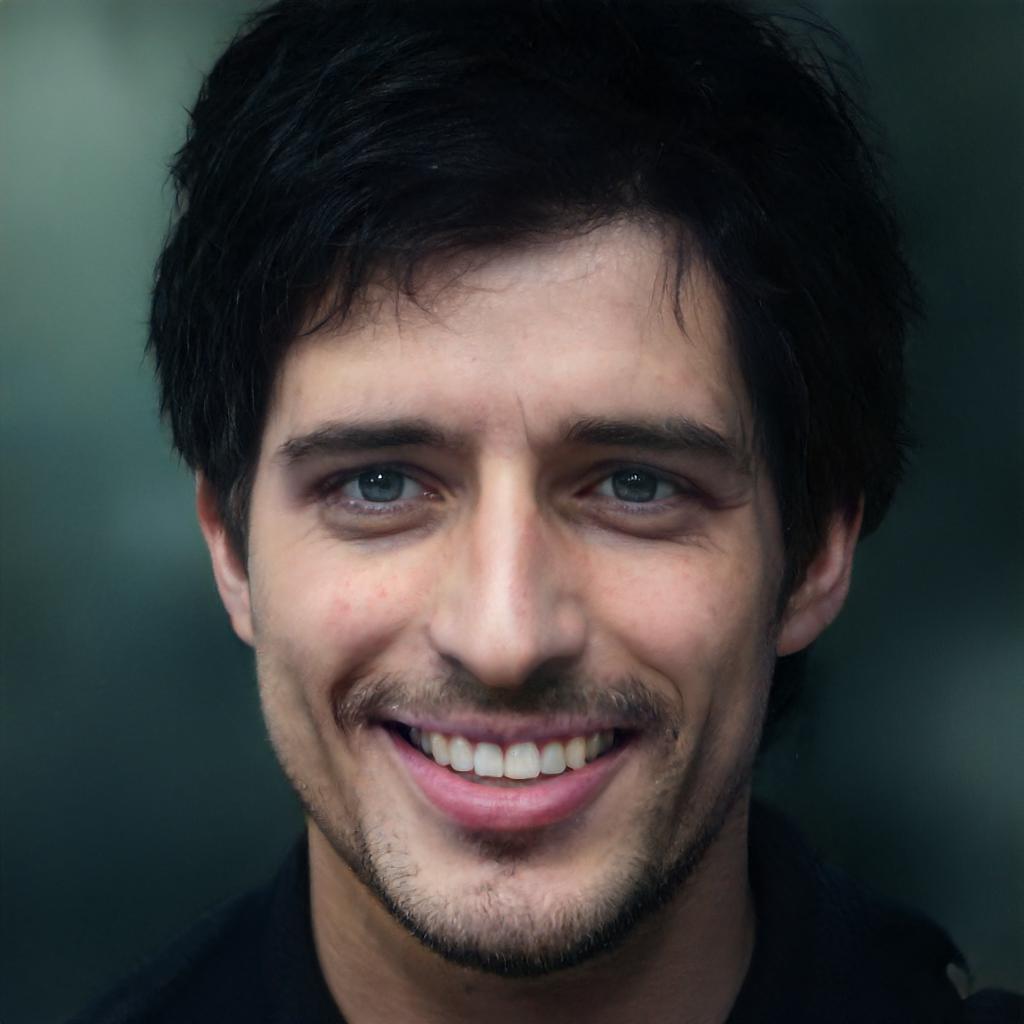} \\
    
    \rotatebox{90}{pSp} &
    \includegraphics[width=.18\linewidth]{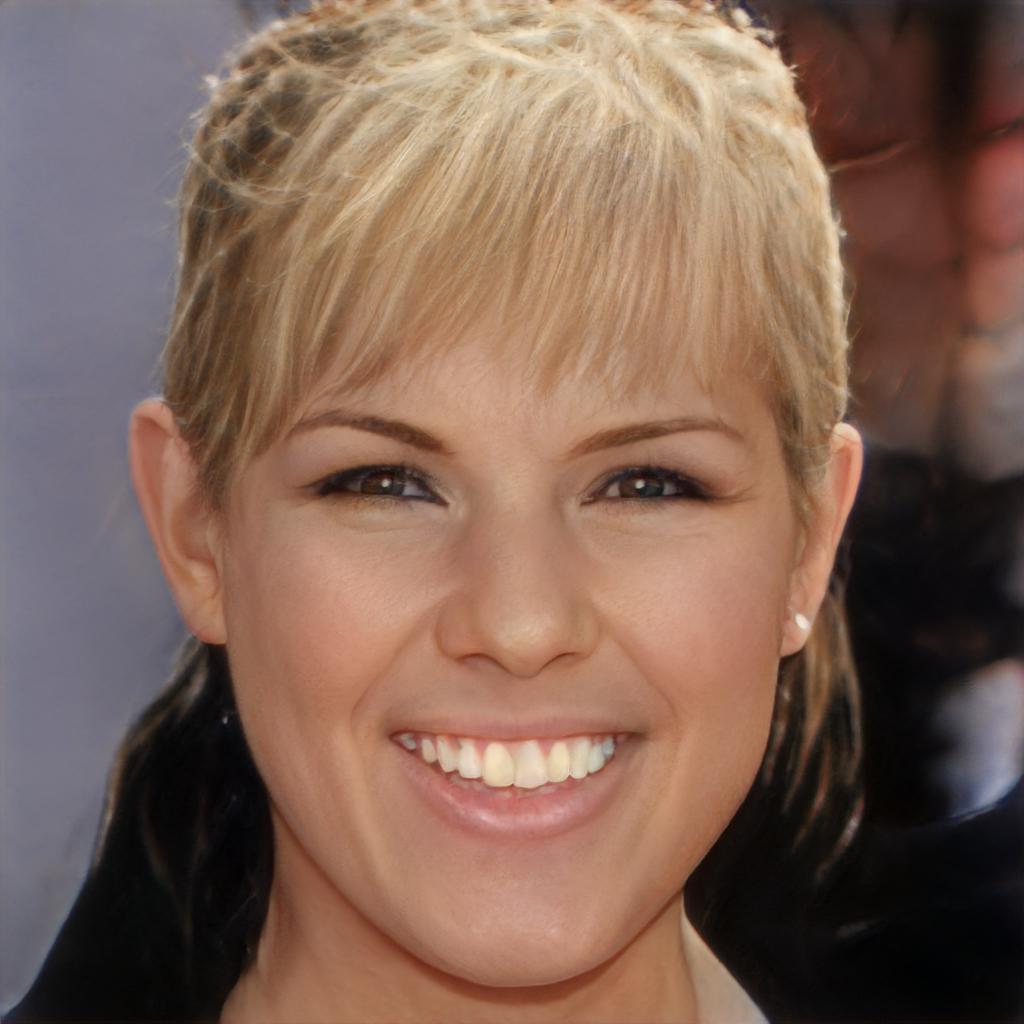} &
    \includegraphics[width=.18\linewidth]{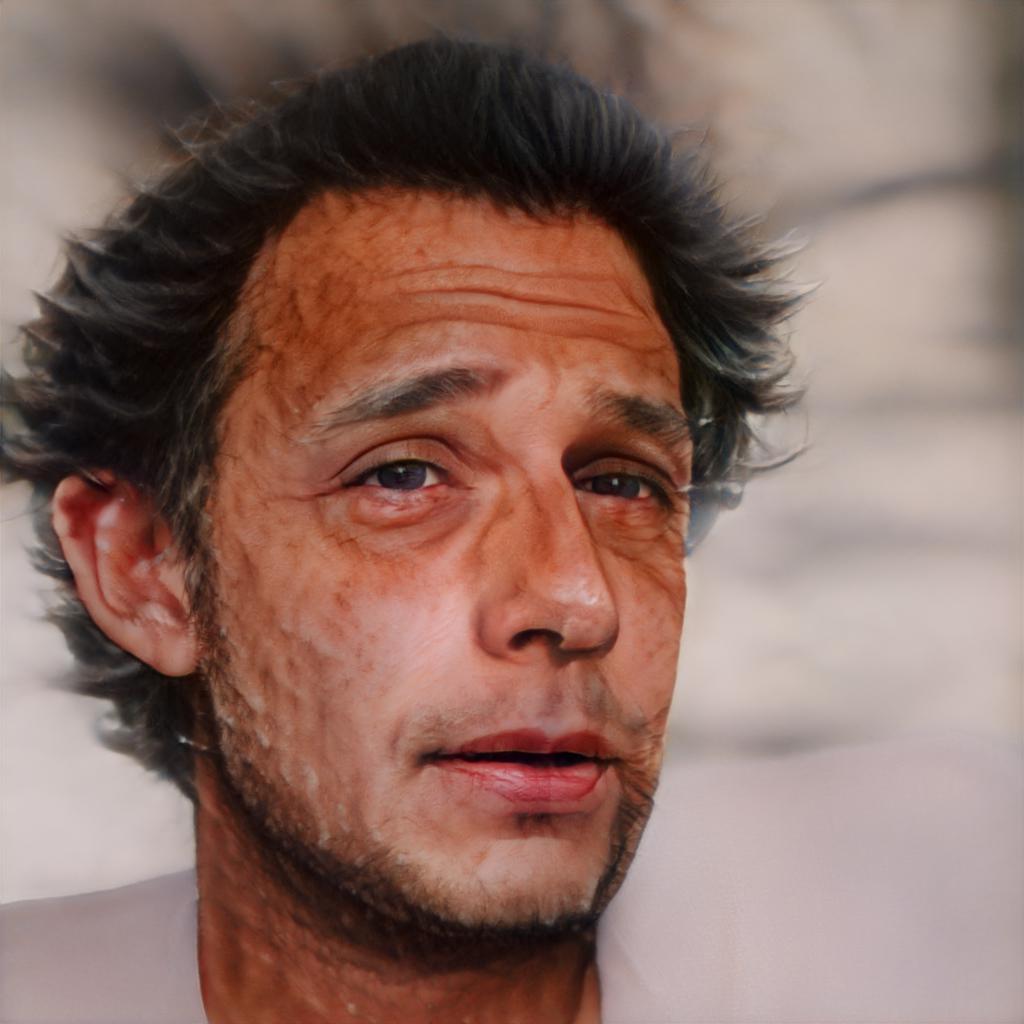} &
    \includegraphics[width=.18\linewidth]{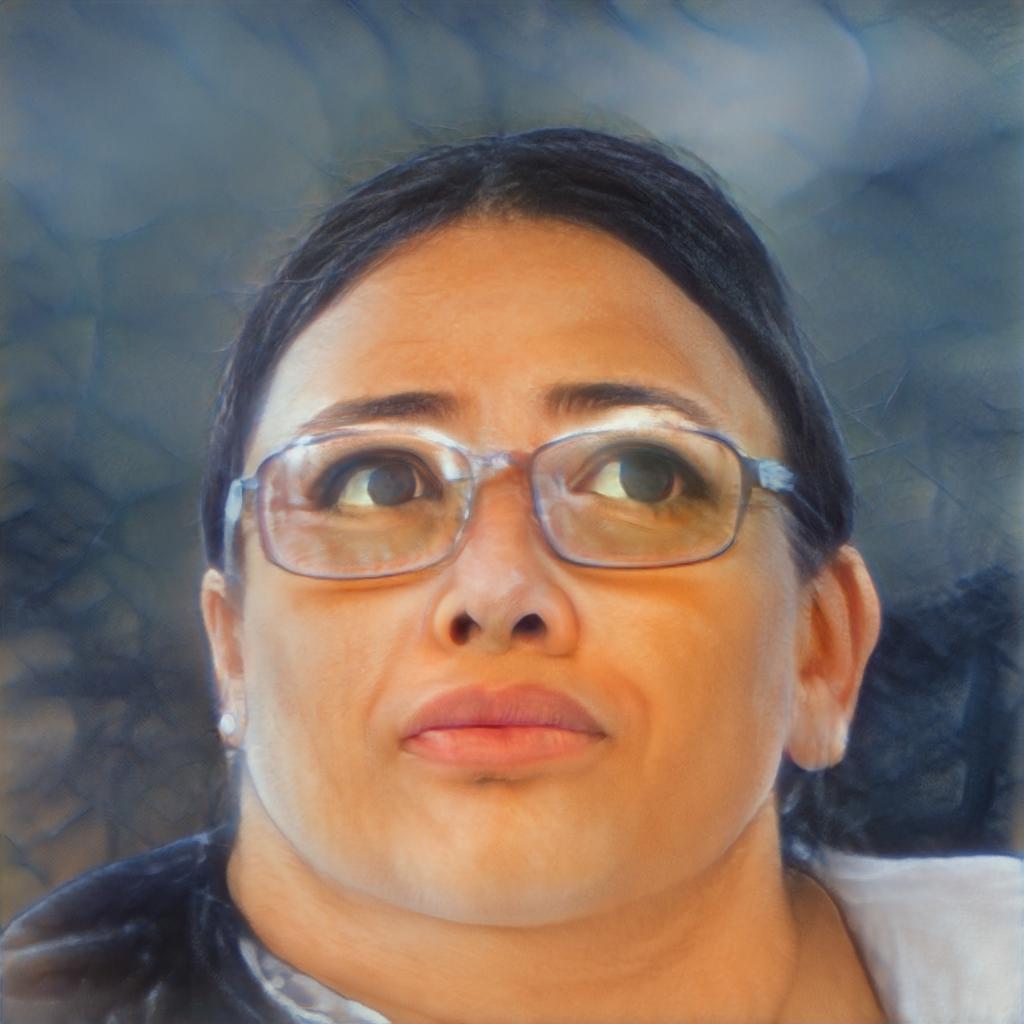} &
    \includegraphics[width=.18\linewidth]{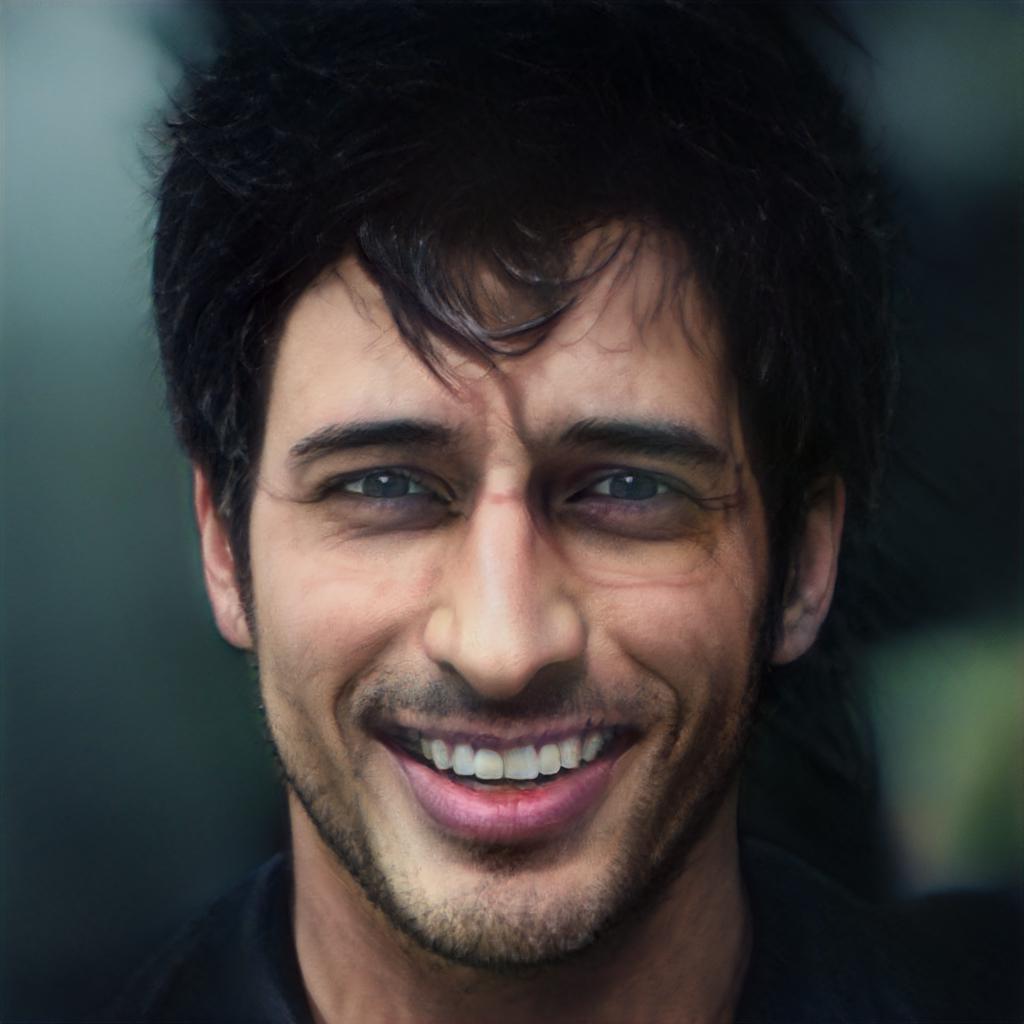} \\
    
    \rotatebox{90}{restyle(e4e)} &
    \includegraphics[width=.18\linewidth]{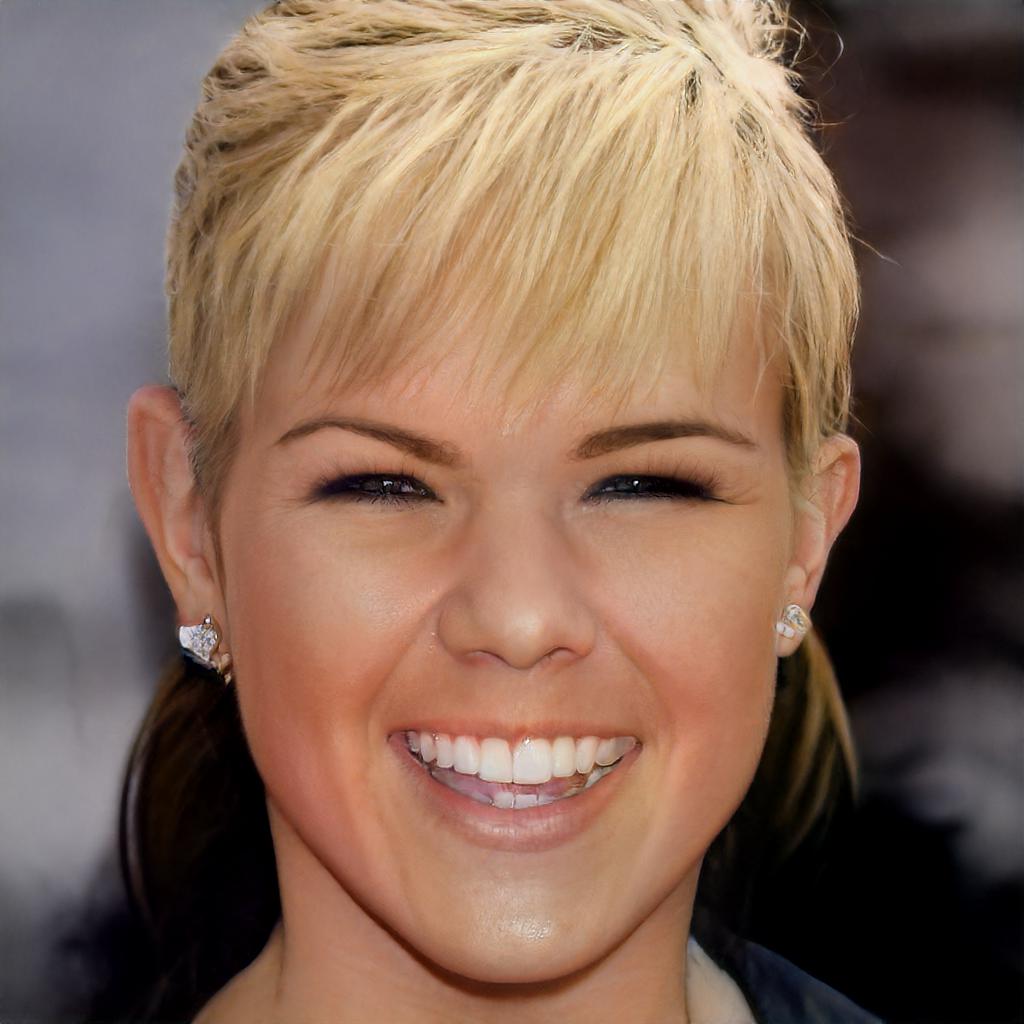} &
    \includegraphics[width=.18\linewidth]{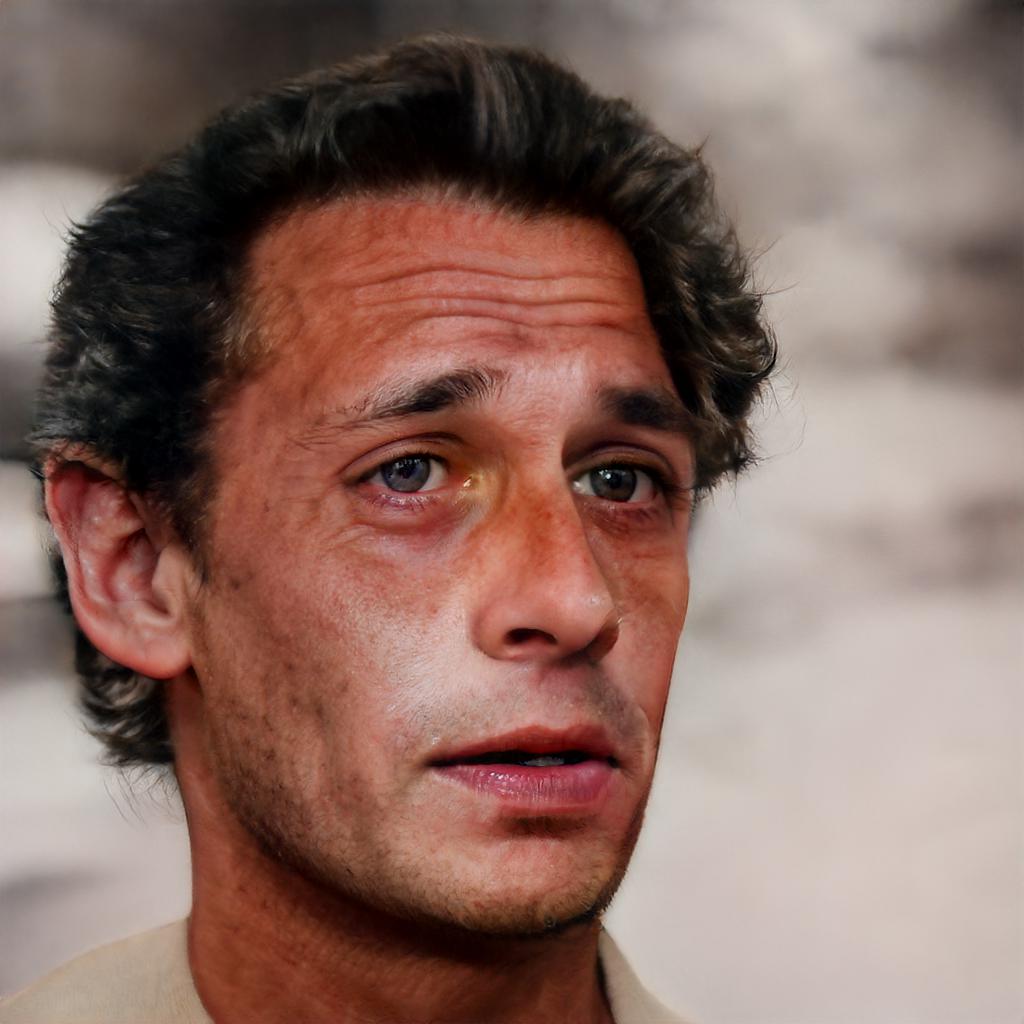} &
    \includegraphics[width=.18\linewidth]{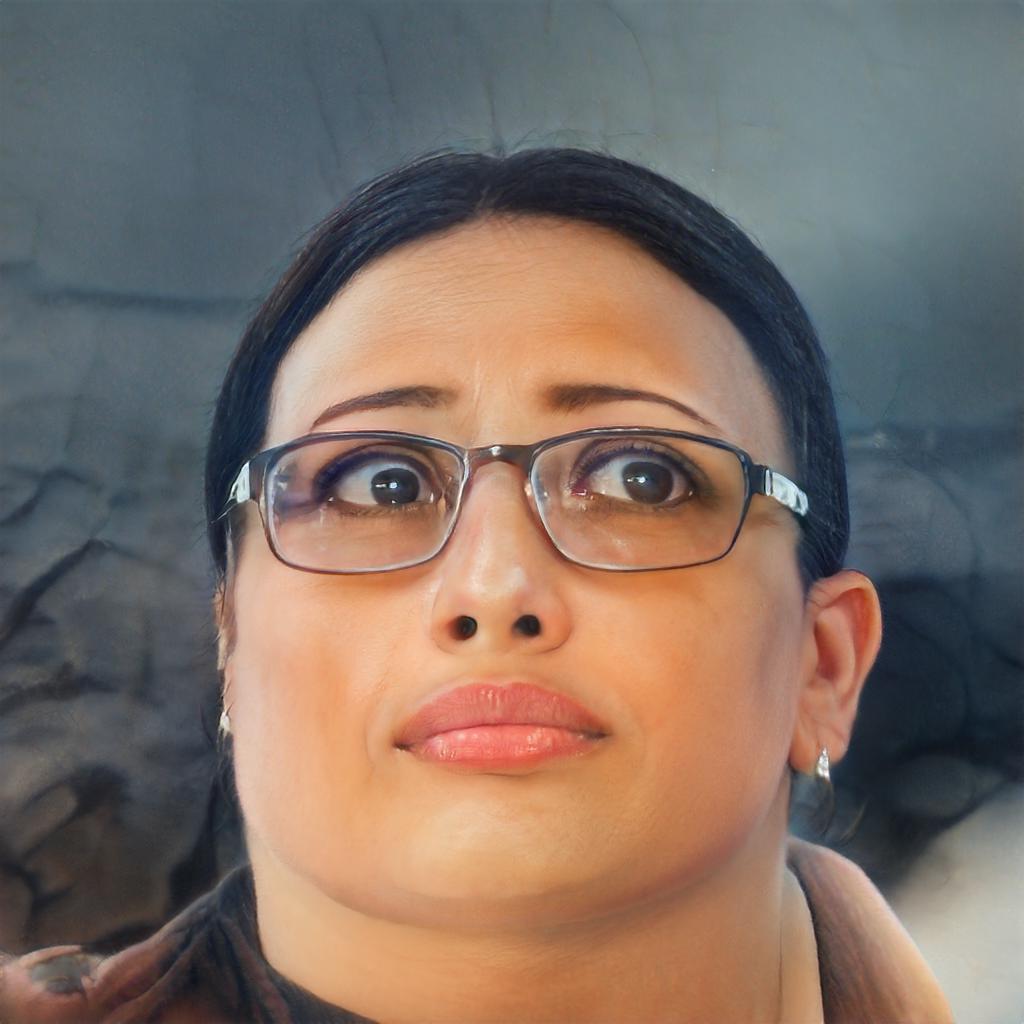} &
    \includegraphics[width=.18\linewidth]{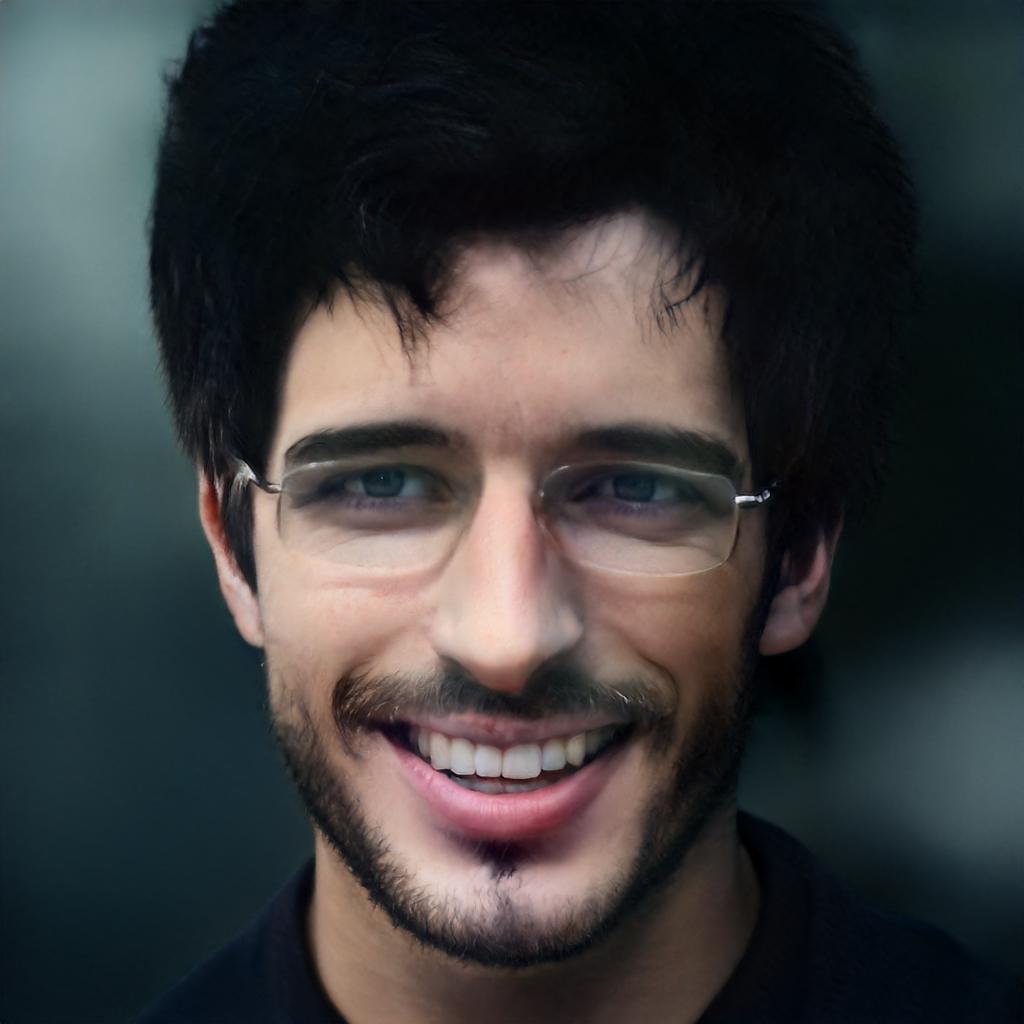} \\
    
    \rotatebox{90}{restyle(pSp)} &
    \includegraphics[width=.18\linewidth]{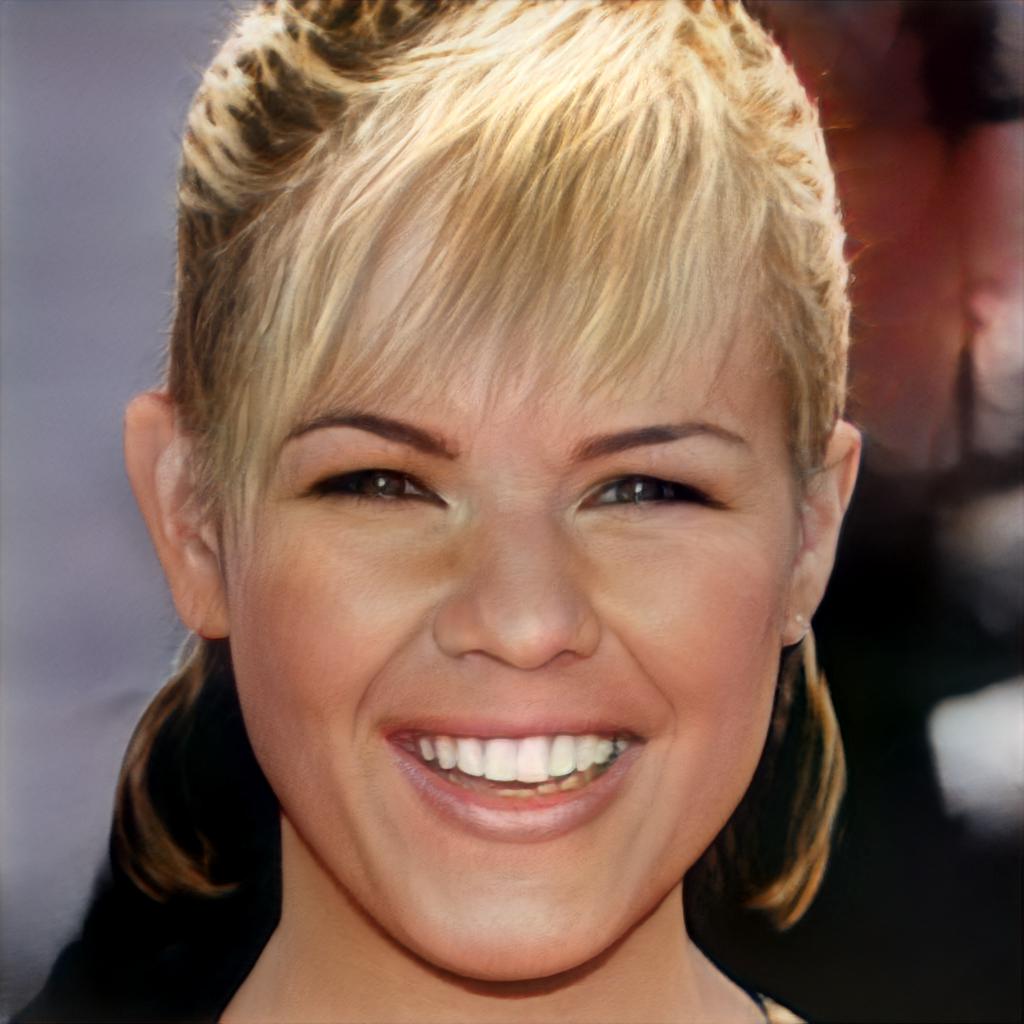} &
    \includegraphics[width=.18\linewidth]{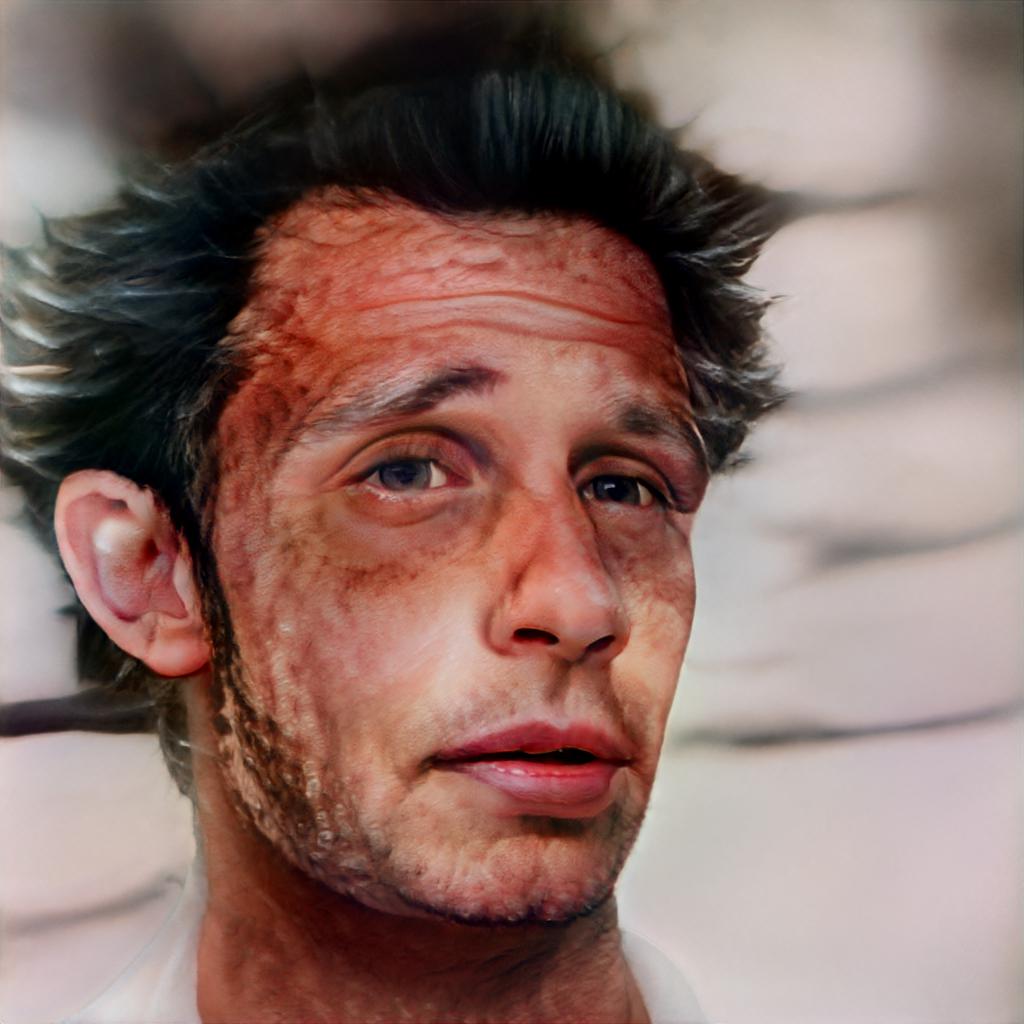} &
    \includegraphics[width=.18\linewidth]{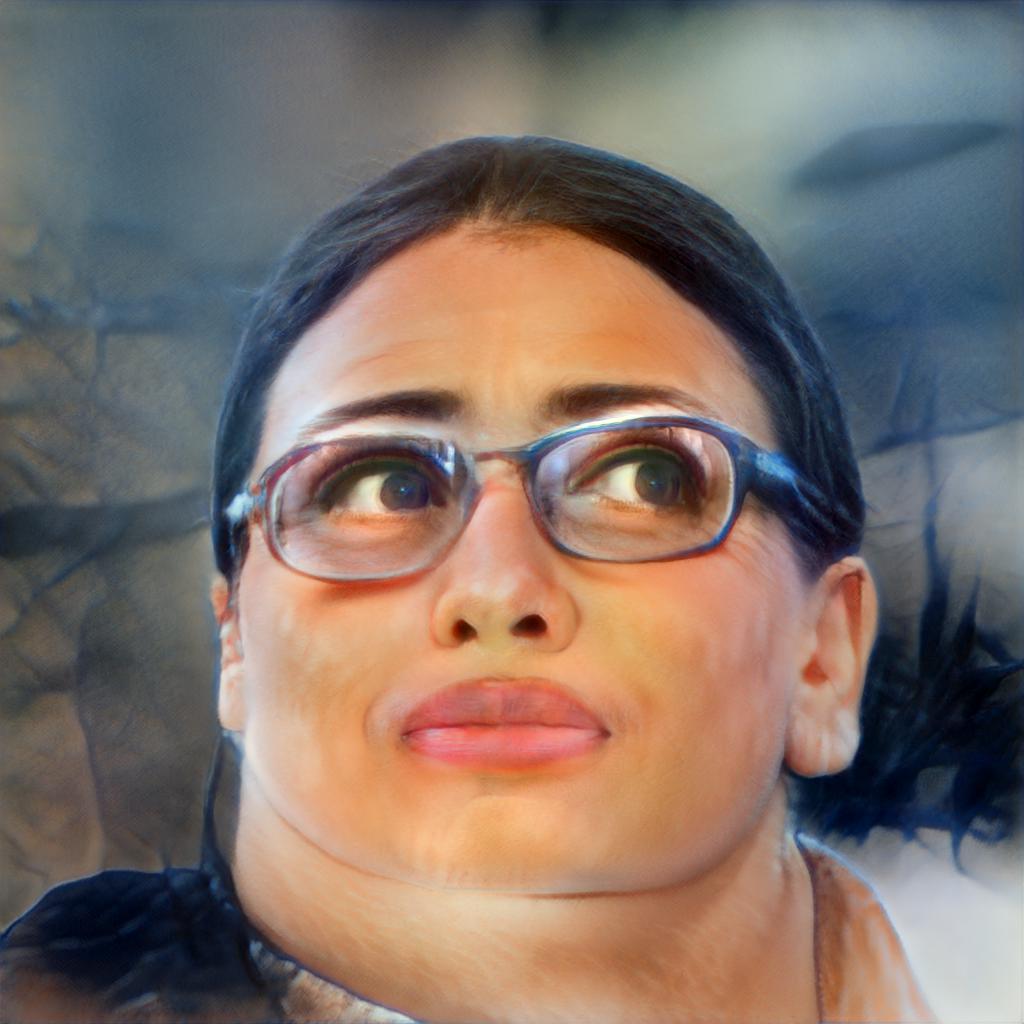} &
    \includegraphics[width=.18\linewidth]{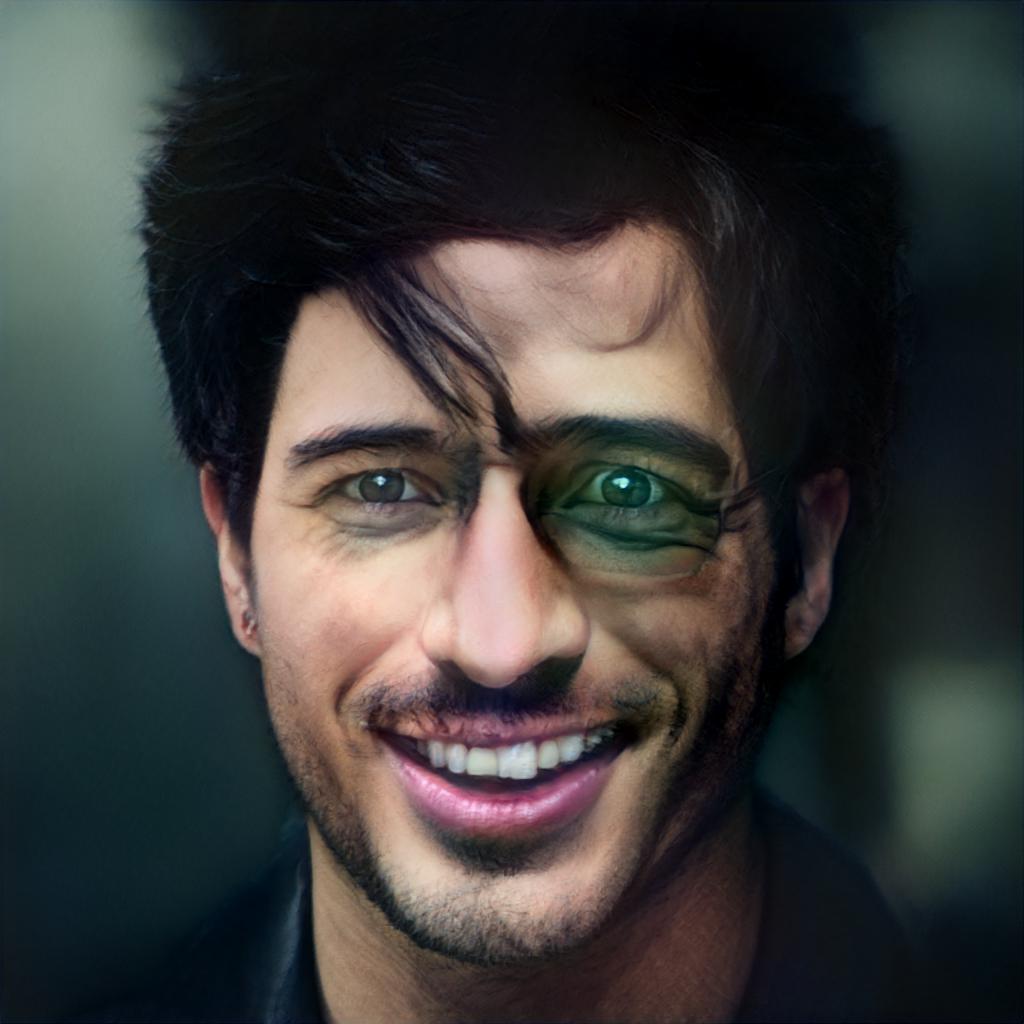} \\
    
    \rotatebox{90}{hfgi} &
    \includegraphics[width=.18\linewidth]{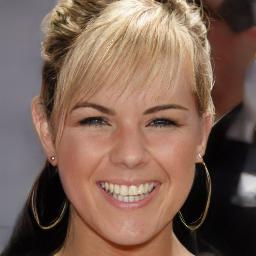} &
    \includegraphics[width=.18\linewidth]{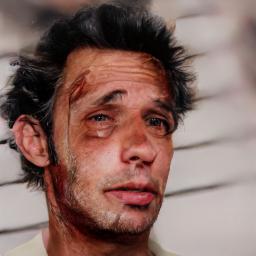} &
    \includegraphics[width=.18\linewidth]{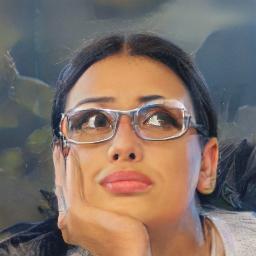} &
    \includegraphics[width=.18\linewidth]{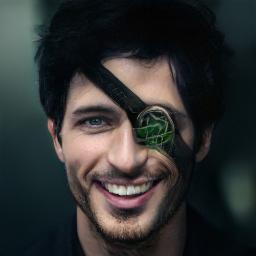} \\
    
    \rotatebox{90}{egain} &
    \includegraphics[width=.18\linewidth]{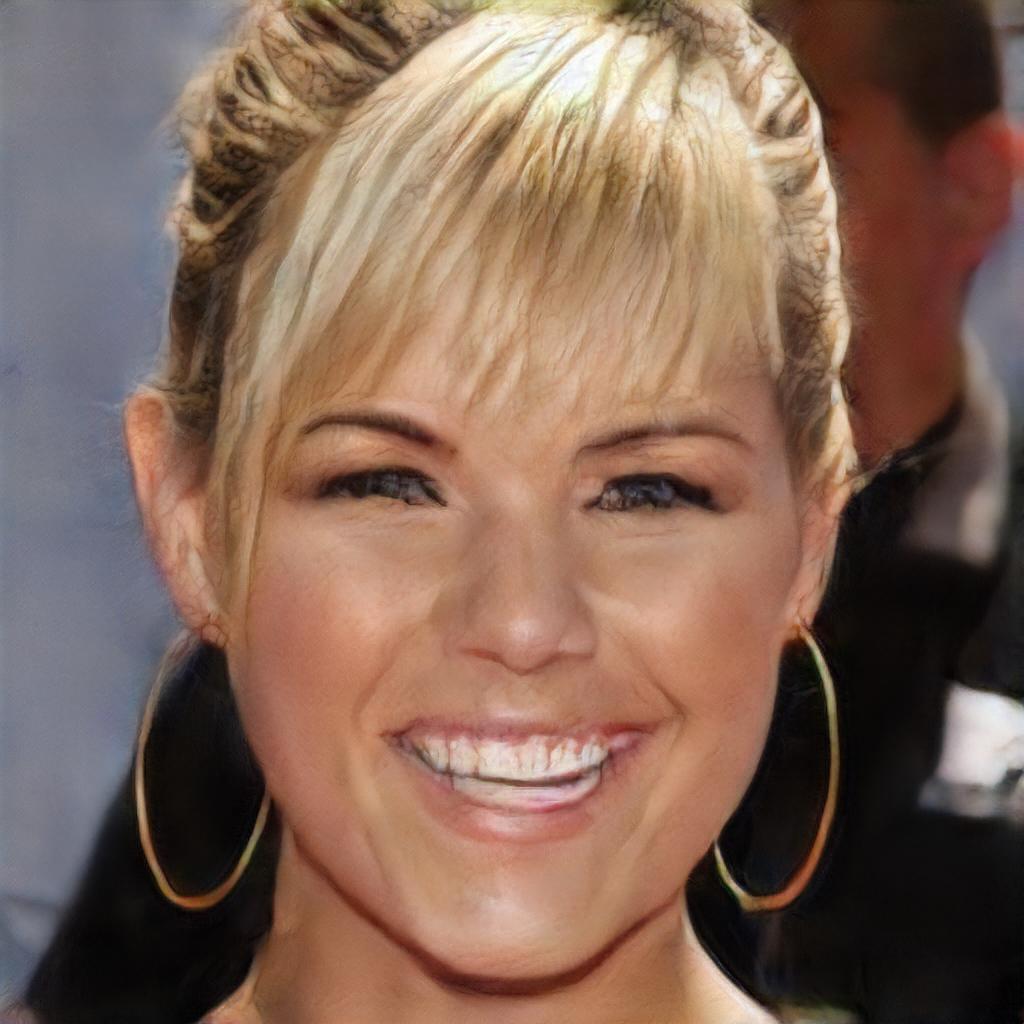} &
    \includegraphics[width=.18\linewidth]{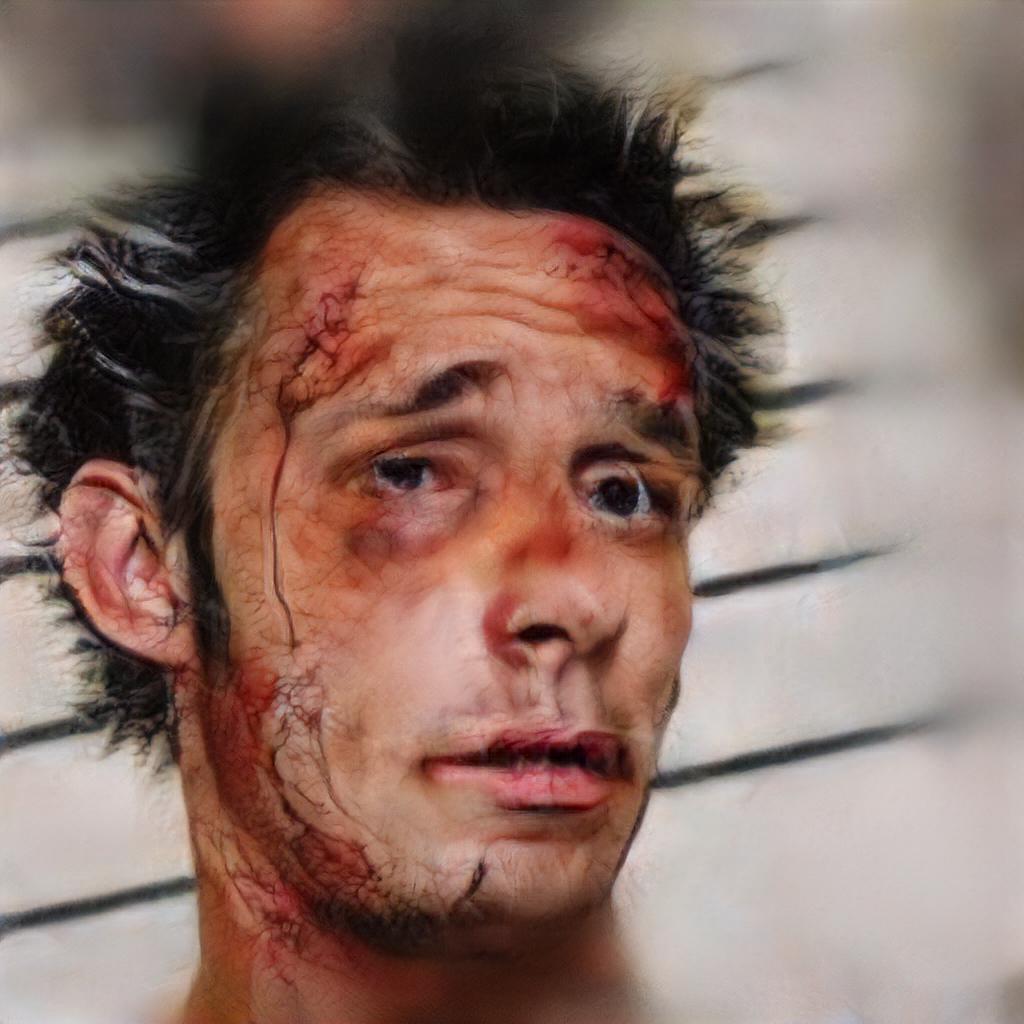} &
    \includegraphics[width=.18\linewidth]{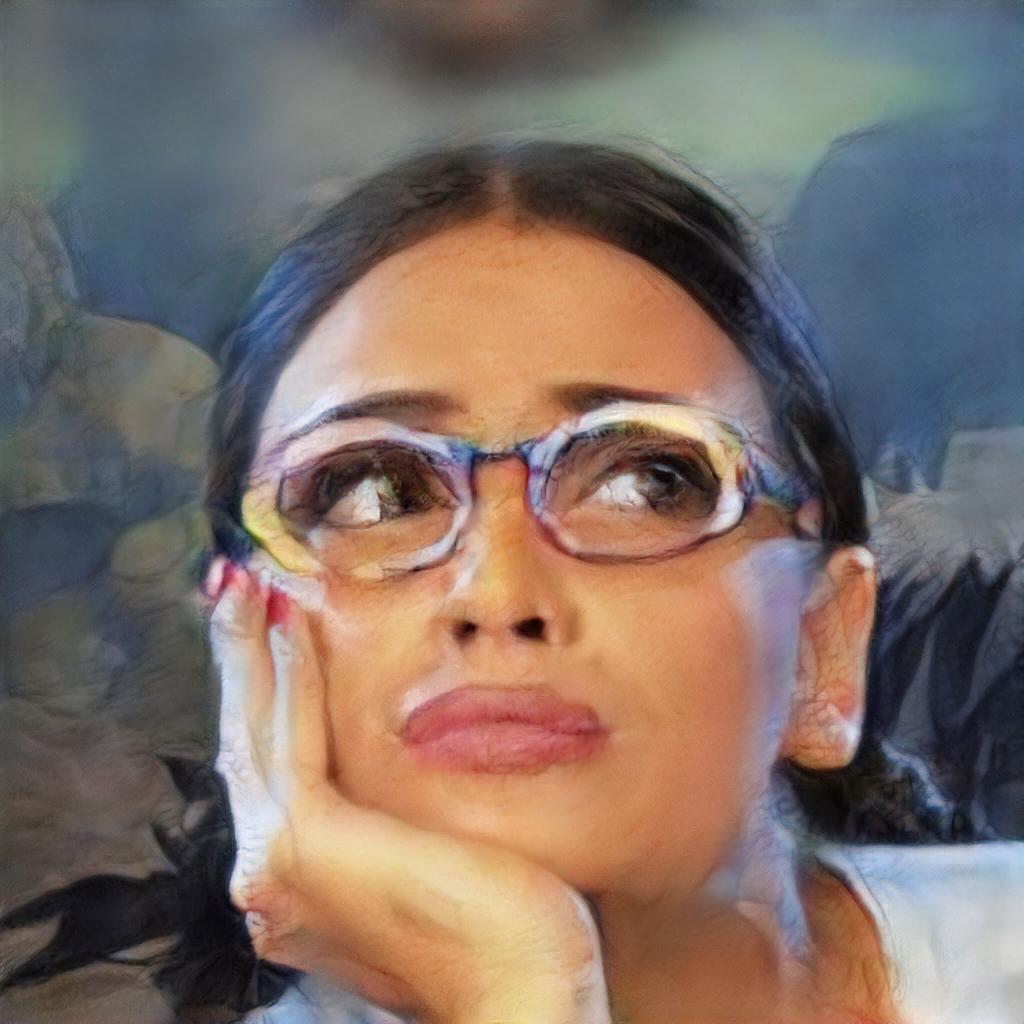} &
    \includegraphics[width=.18\linewidth]{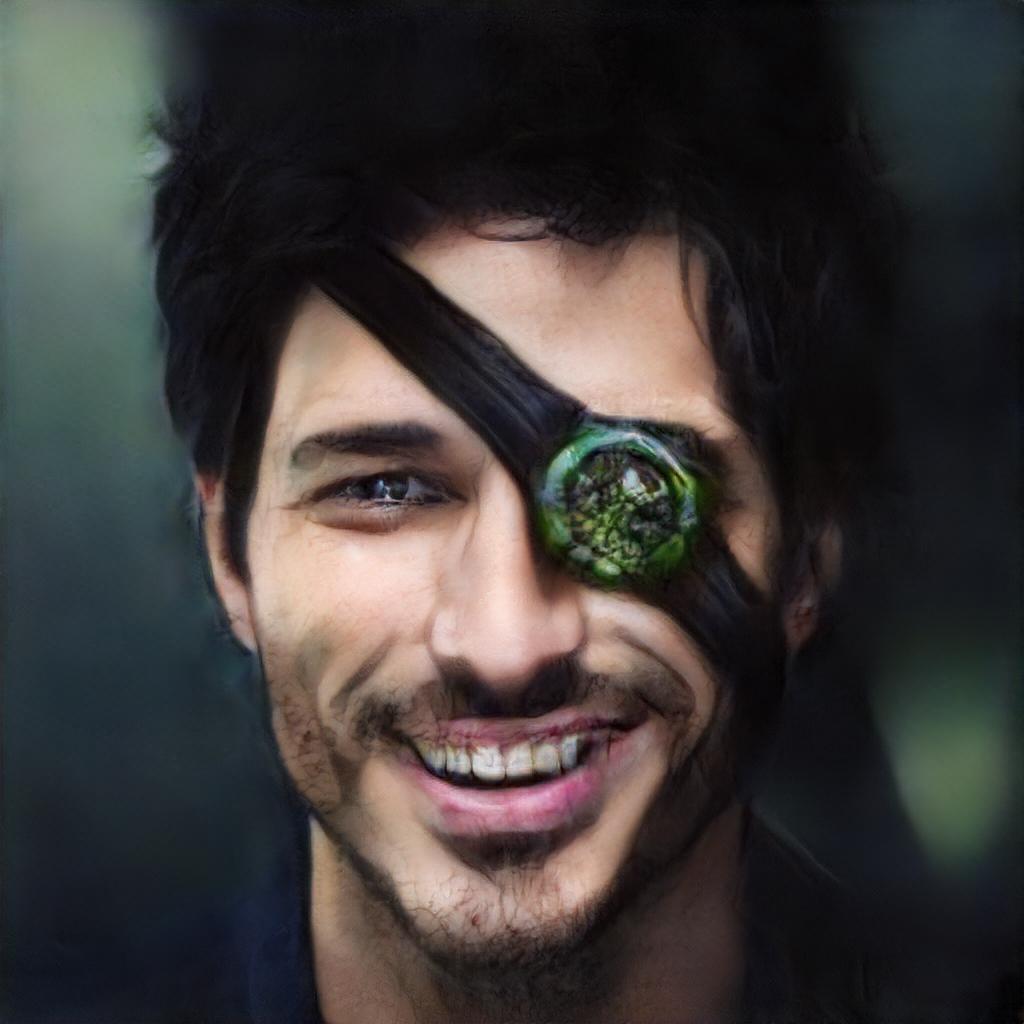} \\
    
\end{tabular}
\caption{Visualisation of reconstruction quality}
\label{fig:qualitative}
\end{figure}

\section{Conclusion and Future Work}
\label{sec:conlusion}

In this paper, we introduce an extended GAN inversion architecture that can be used as a reference when constructing GAN inversion models. It addresses some of the inherent challenges in GAN inversion such as the trade-off between latent code editability and reconstructed image quality. This is done by introducing dedicated components to capture the lost image details and reincorporate them in the reconstructed image. We propose a specific model \textit{egain} which adopts the design suggestions in EGAIN and uses SWAGAN as the generator module. The experimental results demonstrate the superior quality of the reconstructed images produced by \textit{egain} over all the state-of-the-art models. Future work will focus on semantic attribute editing in the reconstructed facial images and how it can be used in the task of evaluating the performance of face recognition systems.

\printbibliography

@misc{radford2016unsupervised,
	title        = {Unsupervised Representation Learning with Deep Convolutional Generative Adversarial Networks},
	author       = {Alec Radford and Luke Metz and Soumith Chintala},
	year         = 2016,
	eprint       = {1511.06434},
	archiveprefix = {arXiv},
	primaryclass = {cs.LG}
}

@misc{mirza2014conditional,
	title        = {Conditional Generative Adversarial Nets},
	author       = {Mehdi Mirza and Simon Osindero},
	year         = 2014,
	eprint       = {1411.1784},
	archiveprefix = {arXiv},
	primaryclass = {cs.LG}
}

@misc{karras2018progressive,
	title        = {Progressive Growing of GANs for Improved Quality, Stability, and Variation},
	author       = {Tero Karras and Timo Aila and Samuli Laine and Jaakko Lehtinen},
	year         = 2018,
	eprint       = {1710.10196},
	archiveprefix = {arXiv},
	primaryclass = {cs.NE}
}

@misc{karras2019stylebased,
	title        = {A Style-Based Generator Architecture for Generative Adversarial Networks},
	author       = {Tero Karras and Samuli Laine and Timo Aila},
	year         = 2019,
	eprint       = {1812.04948},
	archiveprefix = {arXiv},
	primaryclass = {cs.NE}
}

@misc{karras2020analyzing,
	title        = {Analyzing and Improving the Image Quality of StyleGAN},
	author       = {Tero Karras and Samuli Laine and Miika Aittala and Janne Hellsten and Jaakko Lehtinen and Timo Aila},
	year         = 2020,
	eprint       = {1912.04958},
	archiveprefix = {arXiv},
	primaryclass = {cs.CV}
}

@misc{huang2017arbitrary,
	title        = {Arbitrary Style Transfer in Real-time with Adaptive Instance Normalization},
	author       = {Xun Huang and Serge Belongie},
	year         = 2017,
	eprint       = {1703.06868},
	archiveprefix = {arXiv},
	primaryclass = {cs.CV}
}

@misc{gal2021swagan,
	title        = {SWAGAN: A Style-based Wavelet-driven Generative Model},
	author       = {Rinon Gal and Dana Cohen and Amit Bermano and Daniel Cohen-Or},
	year         = 2021,
	eprint       = {2102.06108},
	archiveprefix = {arXiv},
	primaryclass = {cs.CV}
}

@misc{xia2021gan,
	title        = {GAN Inversion: A Survey},
	author       = {Weihao Xia and Yulun Zhang and Yujiu Yang and Jing-Hao Xue and Bolei Zhou and Ming-Hsuan Yang},
	year         = 2021,
	eprint       = {2101.05278},
	archiveprefix = {arXiv},
	primaryclass = {cs.CV}
}

@misc{zhu2018generative,
	title        = {Generative Visual Manipulation on the Natural Image Manifold},
	author       = {Jun-Yan Zhu and Philipp Krähenbühl and Eli Shechtman and Alexei A. Efros},
	year         = 2018,
	eprint       = {1609.03552},
	archiveprefix = {arXiv},
	primaryclass = {cs.CV}
}

@misc{pan2020exploiting,
	title        = {Exploiting Deep Generative Prior for Versatile Image Restoration and Manipulation},
	author       = {Xingang Pan and Xiaohang Zhan and Bo Dai and Dahua Lin and Chen Change Loy and Ping Luo},
	year         = 2020,
	eprint       = {2003.13659},
	archiveprefix = {arXiv},
	primaryclass = {eess.IV}
}

@misc{abdal2019image2stylegan,
	title        = {Image2StyleGAN: How to Embed Images Into the StyleGAN Latent Space?},
	author       = {Rameen Abdal and Yipeng Qin and Peter Wonka},
	year         = 2019,
	eprint       = {1904.03189},
	archiveprefix = {arXiv},
	primaryclass = {cs.CV}
}

@inproceedings{alaluf2021restyle,
	title        = {ReStyle: A Residual-Based StyleGAN Encoder via Iterative Refinement},
	author       = {Alaluf, Yuval and Patashnik, Or and Cohen-Or, Daniel},
	year         = 2021,
	month        = {10},
	booktitle    = {Proceedings of the IEEE/CVF International Conference on Computer Vision (ICCV)}
}

@inproceedings{richardson2021encoding,
	title        = {Encoding in Style: a StyleGAN Encoder for Image-to-Image Translation},
	author       = {Richardson, Elad and Alaluf, Yuval and Patashnik, Or and Nitzan, Yotam and Azar, Yaniv and Shapiro, Stav and Cohen-Or, Daniel},
	year         = 2021,
	month        = {6},
	booktitle    = {IEEE/CVF Conference on Computer Vision and Pattern Recognition}
}

@article{tov2021designing,
	title        = {Designing an Encoder for StyleGAN Image Manipulation},
	author       = {Tov, Omer and Alaluf, Yuval and Nitzan, Yotam and Patashnik, Or and Cohen-Or, Daniel},
	year         = 2021,
	journal      = {arXiv preprint arXiv:2102.02766}
}

@article{wang2021HFGI,
	title        = {High-Fidelity GAN Inversion for Image Attribute Editing},
	author       = {Tengfei Wang and Yong Zhang and Yanbo Fan and Jue Wang and Qifeng Chen},
	year         = 2021,
	journal      = {arxiv:2109.06590}
}

@article{wang2004ssim,
	title        = {Image quality assessment: from error visibility to structural similarity},
	author       = {Zhou Wang and Bovik, A.C. and Sheikh, H.R. and Simoncelli, E.P.},
	year         = 2004,
	journal      = {IEEE Transactions on Image Processing},
	volume       = 13,
	number       = 4,
	pages        = {600--612},
	doi          = {10.1109/TIP.2003.819861}
}

@misc{zhu2020indomain,
	title        = {In-Domain GAN Inversion for Real Image Editing},
	author       = {Jiapeng Zhu and Yujun Shen and Deli Zhao and Bolei Zhou},
	year         = 2020,
	eprint       = {2004.00049},
	archiveprefix = {arXiv},
	primaryclass = {cs.CV}
}

@misc{pidhorskyi2020adversarial,
	title        = {Adversarial Latent Autoencoders},
	author       = {Stanislav Pidhorskyi and Donald Adjeroh and Gianfranco Doretto},
	year         = 2020,
	eprint       = {2004.04467},
	archiveprefix = {arXiv},
	primaryclass = {cs.LG}
}

@misc{tishby2015deep,
	title        = {Deep Learning and the Information Bottleneck Principle},
	author       = {Naftali Tishby and Noga Zaslavsky},
	year         = 2015,
	eprint       = {1503.02406},
	archiveprefix = {arXiv},
	primaryclass = {cs.LG}
}

@misc{pinnimty2020transforming,
	title        = {Transforming Facial Weight of Real Images by Editing Latent Space of StyleGAN},
	author       = {V N S Rama Krishna Pinnimty and Matt Zhao and Palakorn Achananuparp and Ee-Peng Lim},
	year         = 2020,
	eprint       = {2011.02606},
	archiveprefix = {arXiv},
	primaryclass = {cs.CV}
}

@misc{chai2021ensembling,
	title        = {Ensembling with Deep Generative Views},
	author       = {Lucy Chai and Jun-Yan Zhu and Eli Shechtman and Phillip Isola and Richard Zhang},
	year         = 2021,
	eprint       = {2104.14551},
	archiveprefix = {arXiv},
	primaryclass = {cs.CV}
}

@misc{deng2019arcface,
	title        = {ArcFace: Additive Angular Margin Loss for Deep Face Recognition},
	author       = {Jiankang Deng and Jia Guo and Niannan Xue and Stefanos Zafeiriou},
	year         = 2019,
	eprint       = {1801.07698},
	archiveprefix = {arXiv},
	primaryclass = {cs.CV}
}

@misc{chen2020ssdgan,
	title        = {SSD-GAN: Measuring the Realness in the Spatial and Spectral Domains},
	author       = {Yuanqi Chen and Ge Li and Cece Jin and Shan Liu and Thomas Li},
	year         = 2020,
	eprint       = {2012.05535},
	archiveprefix = {arXiv},
	primaryclass = {cs.CV}
}

@misc{durall2020watch,
	title        = {Watch your Up-Convolution: CNN Based Generative Deep Neural Networks are Failing to Reproduce Spectral Distributions},
	author       = {Ricard Durall and Margret Keuper and Janis Keuper},
	year         = 2020,
	eprint       = {2003.01826},
	archiveprefix = {arXiv},
	primaryclass = {cs.CV}
}

@misc{dzanic2020fourier,
	title        = {Fourier Spectrum Discrepancies in Deep Network Generated Images},
	author       = {Tarik Dzanic and Karan Shah and Freddie Witherden},
	year         = 2020,
	eprint       = {1911.06465},
	archiveprefix = {arXiv},
	primaryclass = {eess.IV}
}

@inproceedings{lee2020celeba,
	title        = {MaskGAN: Towards Diverse and Interactive Facial Image Manipulation},
	author       = {Lee, Cheng-Han and Liu, Ziwei and Wu, Lingyun and Luo, Ping},
	year         = 2020,
	booktitle    = {IEEE Conference on Computer Vision and Pattern Recognition}
}

@inproceedings{liu2015faceattributes,
	title        = {Deep Learning Face Attributes in the Wild},
	author       = {Liu, Ziwei and Luo, Ping and Wang, Xiaogang and Tang, Xiaoou},
	year         = 2015,
	month        = {12},
	booktitle    = {Proceedings of International Conference on Computer Vision (ICCV)}
}

@inproceedings{meng2021magface,
	title        = {{MagFace}: A universal representation for face recognition and quality assessment},
	author       = {Meng, Qiang and Zhao, Shichao and Huang, Zhida and Zhou, Feng},
	year         = 2021,
	booktitle    = CVPR
}

@article{rowden2018quality,
	title        = {Learning Face Image Quality From Human Assessments},
	author       = {Best-Rowden, Lacey and Jain, Anil K.},
	year         = 2018,
	journal      = {IEEE Transactions on Information Forensics and Security},
	volume       = 13,
	number       = 12,
	pages        = {3064--3077},
	doi          = {10.1109/TIFS.2018.2799585}
}

@article{Sheikh2006vifp,
	title        = {Image information and visual quality},
	author       = {Sheikh, H.R. and Bovik, A.C.},
	year         = 2006,
	journal      = {IEEE Transactions on Image Processing},
	volume       = 15,
	number       = 2,
	pages        = {430--444},
	doi          = {10.1109/TIP.2005.859378}
}

@article{zhou1998scc,
	title        = {A wavelet transform method to merge Landsat TM and SPOT panchromatic data},
	author       = {Zhou, J. and Civco, D. L. and Silander, J. A.},
	year         = 1998,
	journal      = {International Journal of Remote Sensing},
	publisher    = {TAYLOR & FRANCIS LTD},
	volume       = 19,
	number       = 4,
	pages        = {743--757},
	doi          = {10.1080/014311698215973},
	issn         = {13665901, 01431161},
	language     = {eng},
	format       = {article}
}

@misc{goodfellow2014generative,
	title        = {Generative Adversarial Networks},
	author       = {Ian J. Goodfellow and Jean Pouget-Abadie and Mehdi Mirza and Bing Xu and David Warde-Farley and Sherjil Ozair and Aaron Courville and Yoshua Bengio},
	year         = 2014,
	eprint       = {1406.2661},
	archiveprefix = {arXiv},
	primaryclass = {stat.ML}
}

@misc{karras2021aliasfree,
      title={Alias-Free Generative Adversarial Networks}, 
      author={Tero Karras and Miika Aittala and Samuli Laine and Erik Härkönen and Janne Hellsten and Jaakko Lehtinen and Timo Aila},
      year={2021},
      eprint={2106.12423},
      archivePrefix={arXiv},
      primaryClass={cs.CV}
}

@article{marcel2021age,
	title        = {Deep Face Age Progression: A Survey},
	author       = {Grimmer, Marcel and Ramachandra, Raghavendra and Busch, Christoph},
	year         = 2021,
	journal      = {IEEE Access},
	volume       = 9,
	number       = {},
	pages        = {83376--83393},
	doi          = {10.1109/ACCESS.2021.3085835}
}

\end{document}